\documentclass{article}

\PassOptionsToPackage{dvipsnames}{xcolor}
\usepackage{iclr2027_conference,times}

\usepackage{amsmath,amssymb}
\usepackage{siunitx}
\usepackage[dvipsnames]{xcolor}
\usepackage{graphicx}
\usepackage[caption=false,font=footnotesize]{subfig}
\usepackage{booktabs}
\usepackage{tikz}
\usepackage{threeparttable}
\usepackage{makecell}
\usepackage{multirow}
\usepackage{rotating}
\usepackage{adjustbox}
\usepackage{soul}
\usepackage{pifont}

\newcommand{\pointmapcomparison}[7][4]{%
  \begin{figure}[t]
    \centering
    \def\pointmapcolumns{#1}%
    \ifnum#1=2\relax
      \def\pointmapwidth{0.49\linewidth}%
    \fi
    \subfloat[RIFE]{#3}\hfill
    \subfloat[TimeLens]{#4}%
    \ifnum#1=2\relax\par\smallskip\else\hfill\fi
    \subfloat[VDM-EVFI]{#5}\hfill
    \subfloat[Ours]{#6}
    \vspace{-1ex}
    \caption{#2}
    \vspace{-1ex}
    \label{#7}
  \end{figure}}

\def\cV{\mathcal{V}} %
\def\cE{\mathcal{E}} %

\def\img{I}

\def\conf{\mathbf{C}}
\def\pmap{\mathbf{X}}

\def\feat{\mathbf{F}}

\def\dec{\mathbf{D}}%
\def\enc{\mathbf{E}}%

\newcommand{\unum}[1]{\underline{\num{#1}}}
\newcommand{\bnum}[1]{\bfseries \num{#1}}

\newcommand{\imgbox}[2]{
\begin{tikzpicture}
\node[anchor=south west, inner sep=0] (img)
    {\includegraphics[width=\linewidth]{#1}};
\begin{scope}[x={(img.south east)}, y={(img.north west)}]
\draw[red, line width=0.5pt] #2;
\end{scope}
\end{tikzpicture}
}

\definecolor{light-gray}{gray}{0.6}

\usepackage{pifont}%
\newcommand{\yesmark}{\ding{51}}%
\newcommand{\nomark}{\ding{55}}%

\def\dustr{DUSt3R}
\def\monstr{MonST3R}
\def\alignr{Align3R}
\def\eagr{EAG3R}
\def\interpr{Interp3R}

\usepackage[breaklinks,hidelinks,colorlinks=false,bookmarks=true,bookmarksnumbered=true]{hyperref}
\usepackage{orcidlink}
\usepackage{url}
\usepackage[capitalize,noabbrev]{cleveref}
\usepackage{flafter}
\usepackage{placeins}

\title{Interp3R: Continuous-time 3D Geometry Estimation with Frames and Events}

\author{
Shuang Guo\textsuperscript{1,2}\orcidlink{0000-0002-0142-0678}
\; Filbert Febryanto\textsuperscript{1,2}
\; Lei Sun\textsuperscript{3}\orcidlink{0000-0001-7310-5565}
\; Luc Van Gool\textsuperscript{3}\orcidlink{0000-0002-3445-5711}
\; Guillermo Gallego\textsuperscript{1,2,4}\orcidlink{0000-0002-2672-9241}
\\
\textsuperscript{1}TU Berlin, Berlin, Germany
\quad \textsuperscript{2}Robotics Institute Germany, Berlin, Germany
\\
\textsuperscript{3}INSAIT, Sofia University ``St. Kliment Ohridski'', Sofia, Bulgaria
\\
\textsuperscript{4}Einstein Center Digital Future and SCIoI Excellence Cluster, Berlin, Germany
}

\iclrfinalcopy

\begin{document}
\maketitle
\fancyhead{}
\renewcommand{\headrulewidth}{0pt}

\begin{abstract}

In recent years, 3D visual foundation models, pioneered by pointmap-based approaches such as \dustr, have attracted a lot of interest, achieving impressive accuracy and strong generalization across diverse scenes. 
However, these methods are inherently limited to recovering scene geometry only at the discrete time instants when images are captured, leaving the scene evolution during the blind time between consecutive frames largely unexplored.
We introduce \interpr{}, to the best of our knowledge, the first method that enhances pointmap-based models to estimate depth and camera poses at arbitrary time instants. 
It leverages asynchronous event data to interpolate pointmaps produced by frame-based models, enabling temporally continuous geometric representations. 
Depth and camera poses are then jointly recovered by aligning the interpolated pointmaps together with those predicted by the underlying frame-based models into a consistent spatial framework.
We train \interpr{} exclusively on a synthetic dataset, yet demonstrate strong generalization across six datasets, both synthetic and real. 
Compared with the best two-stage baseline, \interpr{} reduces absolute relative depth error by 15\%--32\% on DSEC and absolute trajectory error by up to 51\% on EDS.

\end{abstract}

\section{Introduction}
\label{sec:intro}

\begin{figure}[t]
  \centering
  \includegraphics[width=0.8\textwidth]{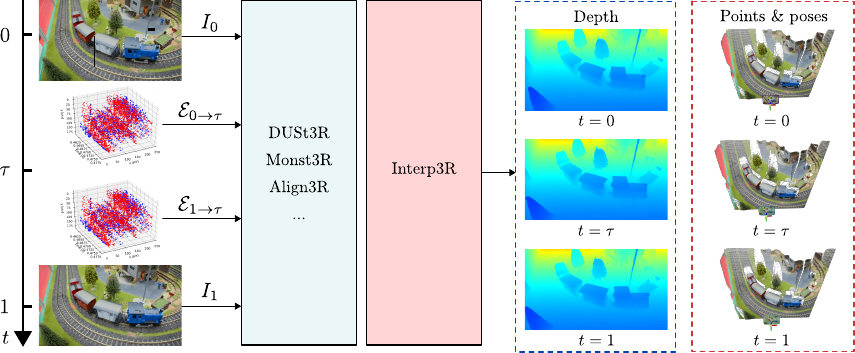}
  \caption{The proposed method takes two images $I_0,I_1$ and the events between them $\cE$ as input (Left), 
  and recovers scene geometry as pointmaps at an arbitrary time instant $\tau$ between the images (Right).
  All available pointmaps can then be used to estimate depth maps and camera poses.
  }
  \label{fig:eyecatcher}
\end{figure}

Recent years have witnessed remarkable progress in 3D visual foundation models. 
Pointmap-based approaches such as DUSt3R \citep{Wang24dus3r} and its successors \citep{Zhang25MonST3R,Lu25Align3R} have demonstrated impressive performance in recovering scene geometry and camera poses from images, while exhibiting strong generalization across diverse environments. 
These models predict dense 3D pointmaps from input frames and enable various downstream tasks, including the estimation of depth and camera poses. 
Despite their success, existing methods fundamentally operate on discrete-time frames and can only recover scene geometry at the times of frame capture.

However, real-world scenes evolve continuously over time. 
The temporal gaps between consecutive frames inevitably introduce blind intervals during which the scene dynamics remain unobserved. 
This limitation becomes particularly severe in scenarios with fast motion or large frame intervals, where significant geometric changes may occur between frames. 
As a result, current frame-based methods fail to provide temporally continuous representations of scene geometry.

Event cameras provide a promising solution to this problem. 
Unlike conventional cameras that capture images at fixed frame rates, event cameras asynchronously record per-pixel brightness changes with microsecond-level temporal resolution \citep{Gallego20pami,Brandli14ssc}.
This allows them to capture fine-grained motion information between frames~\citep{Hidalgo22cvpr,Guo25iccv}, making them well suited for recovering scene evolution in continuous time \citep{Guo24tro,Guo24epba}.
However, effectively integrating asynchronous event streams with pointmap-based models remains challenging. 
Events are sparse, noisy, and fundamentally different from frame-based representations, making their direct use in existing 3D visual models non-trivial.

In this paper, we introduce \emph{\interpr{}}, a novel method that extends pointmap-based models to continuous-time 3D geometry estimation (see \cref{fig:eyecatcher}). 
Given two frames processed by a frame-based model, e.g., \dustr~\citep{Wang24dus3r}, \monstr{}~\citep{Zhang25MonST3R} or \alignr{}~\citep{Lu25Align3R}, \interpr{} leverages the co-captured event stream to interpolate pointmaps at \emph{arbitrary} time instants between the frames. 
The interpolated pointmaps, together with those predicted from the input frames, are then aligned within a unified spatial framework in a coarse-to-fine manner in order to jointly recover depth and camera poses. 
This produces robust, progressively refined geometry estimation in continuous time across large temporal gaps.

We train \interpr{} exclusively on synthetic data, yet the learned model generalizes well to a wide range of synthetic and real-world datasets. 
Extensive experiments demonstrate that \interpr{} consistently outperforms baselines that follow a two-stage pipeline of 2D video frame interpolation followed by frame-based 3D geometry estimation. 
Our results show that incorporating event data enables accurate and temporally continuous reconstruction of camera motion and scene geometry.

Our contributions can be summarized as follows:
\begin{enumerate}
    \item We propose \interpr{}, the \emph{first method} that extends pointmap-based models to estimate camera poses and scene geometry in continuous time, enabling the recovery of depth and camera poses at arbitrary time instants. 
    It leverages asynchronous event data to interpolate pointmaps produced by frame-based models, and jointly recovers geometry by aligning the interpolated and frame-based pointmaps in a coarse-to-fine manner.
    \item We show that \interpr{} is compatible with existing pointmap-based methods (e.g., \monstr{} and \alignr{}), enhancing them to continuous-time space.
    \item We train \interpr{} exclusively on a synthetic dataset and demonstrate strong generalization across six synthetic and real-world datasets. 
    Extensive experiments show that our method outperforms baseline methods, which follow a two-stage pipeline design of 2D video frame interpolation followed by frame-based 3D geometry estimation.
    \item We make the source code publicly available. %
\end{enumerate}

\section{Related Work}
\label{sec:related}

\subsection{3D Geometry Estimation from Images and Videos}

Recent years have witnessed significant progress in estimating scene geometry and camera motion from images.
Here we focus on reviewing the most relevant pointmap-based methods. 
A more detailed tabular comparison is provided in \cref{tab:related} in the Appendix.
\dustr{}~\citep{Wang24dus3r} introduces a pointmap representation and a model that predicts pointmaps from image pairs. 
It recovers scene geometry and camera poses through a global alignment of all predicted pointmaps, and demonstrates strong generalization across diverse scenes. 
Building on this idea, \monstr{}~\citep{Zhang25MonST3R} extends the framework to dynamic scenes by finetuning on dynamic datasets while introducing motion-aware global alignment. 
\alignr{}~\citep{Lu25Align3R} further integrates monocular depth priors into the pointmap framework, combining feed-forward depth models and pointmap-based geometry estimation to improve temporal consistency.

Beyond per-frame geometry estimation, recent work has also explored modeling 3D motion between frames. 
DynaDUSt3R~\citep{Jin25cvpr} augments \dustr{} with motion heads to predict 3D scene flow between two frames at an intermediate query time. 
Given two input images, the model predicts 3D displacement vectors that describe how points move in space, enabling the estimation of point trajectories across time. 
However, while DynaDUSt3R can hallucinate sparse 3D motion between frames, it cannot recover camera poses or dense depth maps at intermediate timestamps

In short, existing 3D geometry estimation methods operate at discrete frame timestamps and lack continuous-time depth and pose estimation models to deal with the blind interval between frames.

\subsection{Event-based Video Frame Interpolation}

Event-based video frame interpolation (EVFI) exploits the high temporal resolution of event cameras to synthesize intermediate frames between sparsely sampled images. 
TimeLens~\citep{Tulyakov21cvpr} and TimeLens++~\citep{Tulyakov22cvpr} pioneer this direction by introducing a framework that estimates inter-frame motion from events and synthesizes intermediate frames accordingly. 
\mbox{REFID}~\citep{Sun23cvpr} performs frame interpolation from blurry video inputs.
\mbox{CBMNet}~\citep{Kim23cvpr} improves motion estimation by jointly leveraging image and event modalities through cross-modal asymmetric bidirectional motion fields. 
TimeLens-XL~\citep{Ma24timelenxl} enables arbitrary-time frame interpolation by recursively decomposing large motions into smaller temporal steps. 
VDM-EVFI~\citep{Chen25cvpr} explores a generative approach by adapting pretrained video diffusion models for event-based frame interpolation, benefiting from large-scale video priors.

Therefore, existing EVFI methods primarily focus on interpolating intermediate frames and remain limited to modeling 2D appearance. 
In contrast, our work directly interpolates 3D geometry in continuous time, recovering both scene structure and camera motion at arbitrary time instants.

\subsection{Event-aided/-based 3D Models}
\label{sec:related:event_aided}

Beyond frame-based 3D geometry estimation, several recent works explore event cameras either as a complementary sensing modality or as a standalone input for robust 3D perception.
Thanks to their high temporal resolution and high dynamic range, event cameras provide rich motion cues that can benefit geometric estimation under fast motion and challenging imaging conditions.

ADAE~\citep{Peng26adaptingdepth} fuses events with monocular depth models~\citep{Yang24cvpr,Yang24neurips} to improve robustness under extreme illumination and motion blur, while retaining per-frame depth prediction.
EAG3R~\citep{Wu25eag3r} augments \monstr{} with event-derived motion cues for robust depth and pose estimation under extreme lighting and fast motion, with predictions tied to discrete frame timestamps.
EventVGGT~\citep{Ren26eventvggt} distills VGGT's geometric priors into event representations for temporally consistent depth estimation.
Its predictions correspond to fixed temporal windows rather than arbitrary continuous query times.
Event3R~\citep{Huang26event3r} explores event-only 3D reconstruction by aggregating spatio-temporal event features, but its geometric predictions are still tied to a fixed reference time rather than arbitrary continuous-time queries.

In summary, existing event-aided and event-based 3D models show that event cameras provide valuable motion information for 3D geometry estimation.
However, they either enhance geometric predictions at discrete frame timestamps or aggregate events in a temporal window to predict geometry at a fixed reference time.
None of them explicitly models geometry in continuous-time space or recovers depth and ego-motion at arbitrary timestamps between the input frames.
In contrast, our work explicitly leverages the quasi-continuous event stream together with frame-based pointmaps to interpolate geometric representations at arbitrary query times, enabling continuous-time depth and camera pose estimation for the first time (see \cref{tab:related} in the Appendix for a detailed comparison).

\section{Method}
\label{sec:method}

\begin{figure}[!t]
  \centering
  \includegraphics[width=\textwidth]{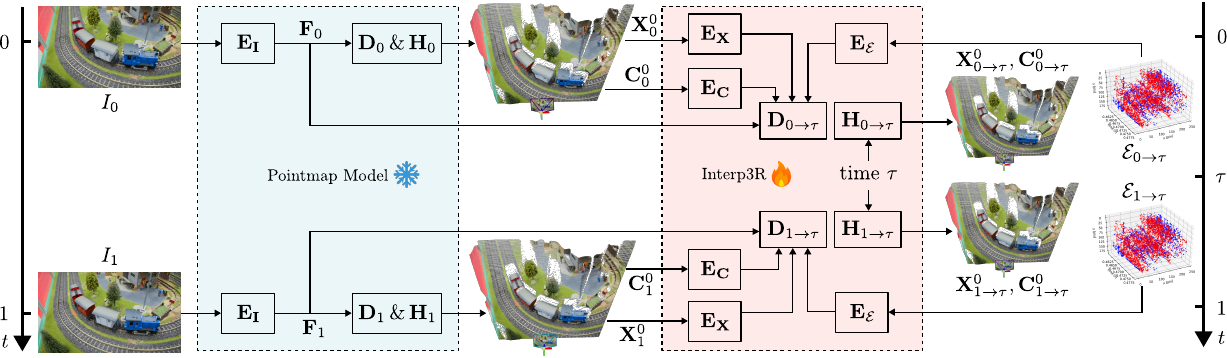}
  \caption{Pipeline for pointmap prediction and interpolation. 
  The pointmap model first takes as input two frames and outputs the source pointmaps. 
  \interpr{} then interpolates the pointmap at the arbitrary target time $\tau \in (0, 1)$ from two directions, with the first part ($0 \rightarrow \tau$, ``forward'') and the second part ($1 \rightarrow \tau$, ``backward'') event data.
  }
  \label{fig:pipeline}
\end{figure}

\subsection{Preliminaries}
\label{sec:method:preliminary}
Our model achieves continuous-time estimation of depth and camera poses by interpolating pointmaps from frame-based models like \dustr{}.
Let us begin with a brief review of \dustr{}, as well as its follow-ups \monstr{} and \alignr, which extend it from static to dynamic scenes.

\textbf{Pointmap-based Geometry Estimation.}
Recent advances in pose-free 3D reconstruction adopt pointmap representations to directly regress dense geometry from image pairs.
DUSt3R predicts per-pixel 3D coordinates (pointmaps) for an image pair $(I_0, I_1)$ using a transformer-based model:
\begin{equation}
\left( \pmap_0^0, \pmap_1^0 \right)
= \mathcal{F}_{\text{DUSt3R}}(I_0, I_1),
\end{equation}
where each pointmap assigns a 3D point to every pixel in the reference view. 
$\pmap_0^0$ and $\pmap_1^0$ indicate the pointmaps of $\img_0$ and $\img_1$ respectively, and both are expressed in the camera coordinate of $\img_0$.
In general, $\pmap_m^n$ indicates the pointmap of $\img_m$ expressed in the camera coordinate of $\img_n$.
These pairwise pointmaps encode 3D scene geometry and can be aligned via a global optimization to recover consistent depth and camera poses across multiple views.

\textbf{Extension to Dynamic Scenes.}
\monstr{} extends DUSt3R to dynamic videos by fine-tuning its pointmap model on dynamic datasets.
Specifically, \monstr{} finds a global state $X^{*}_{\text{global}}$ by minimizing an objective that combines alignment, motion smoothness, and image-based flow constraints:
\begin{equation}
\mathcal{L}_{\text{\monstr{}}}(X^{*}_{\text{global}})
= \mathcal{L}_{\text{align}}
+ w_{\text{smooth}} \mathcal{L}_{\text{smooth}}
+ w_{\text{flow}} \mathcal{L}_{\text{flow}}.
\label{eq:monstr_loss}
\end{equation}
This formulation enables joint estimation of video depth and camera motion in dynamic scenes.

\textbf{Incorporating Monocular Depth Priors.}
Instead of relying solely on RGB images, \alignr{} injects per-frame monocular depth estimates to guide pointmap regression, providing stronger geometric initialization and improving depth accuracy and stability, especially for dynamic scenes.

\begin{figure}[!t]
  \centering
  {\includegraphics[trim={0 0.22cm 0 0.14cm},clip,width=\linewidth]{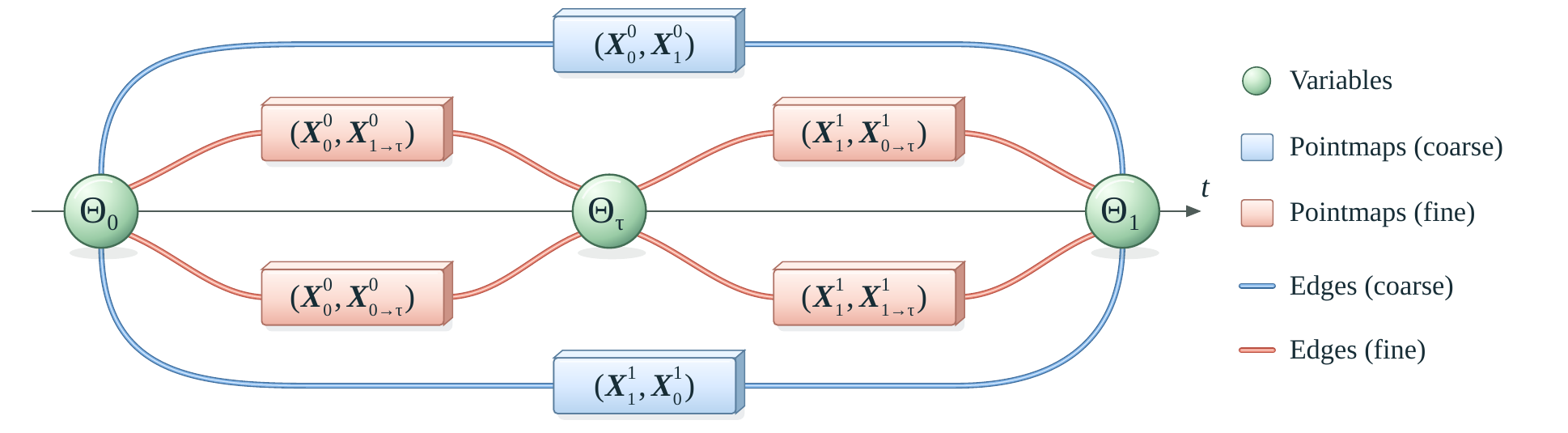}}
  \vspace{-3ex}
  \caption{Coarse-to-fine global alignment.
           Green nodes $\Theta_t$ collect the camera intrinsics, pose and depth variables at time $t$.
           Each $\pmap_m^n$ is a per-pixel 3D pointmap of $I_m$ expressed in $I_n$'s camera coordinate.
           In the coarse alignment (blue), the pointmap pairs connect and constrain only the two source nodes at $t=0$ and $t=1$.
           In the fine alignment (red), the interpolated node at $t=\tau$ is connected to both source nodes through the red pointmap pairs and all node variables are refined.
           }
  \label{fig:alignment}
  \vspace{-2ex}
\end{figure}

\subsection{Pairwise Pointmap Interpolation}
\label{sec:method:interp}
As depicted in \cref{fig:pipeline}, a pointmap model (we use \alignr{} as the main model in this work) takes as input an image pair $(I_0, I_1)$ and outputs the pairwise pointmaps $(\pmap_0^0, \pmap_1^0)$ as well as the corresponding confidence maps $(\conf_0^0, \conf_1^0)$ (i.e., a confidence value per 3D point).

Then, \interpr{} takes as input these pointmap predictions, the RGB features $(\feat_0, \feat_1)$ of the two input frames $(\img_0, \img_1)$ and the bidirectional event data $(\cE_{0 \rightarrow \tau}, \cE_{1 \rightarrow \tau})$, to interpolate the pointmaps $(\pmap_{0 \rightarrow \tau}^0, \pmap_{1 \rightarrow \tau}^0)$ at the target time $\tau \in (0,1)$ from both directions $0 \rightarrow \tau$ and $1 \rightarrow \tau$.
Note that both interpolated pointmaps are expressed in the camera coordinate of $\img_0$.

\textbf{Incorporating Event Data and Pointmaps.}
The events between the two frames are divided into a forward set, $\cE_{0 \rightarrow \tau}$, and a backward set, $\cE_{\tau \rightarrow 1}$.
The backward set $\cE_{\tau \rightarrow 1}$ is formed by reversing the timestamps and polarities of all events, resulting in $\cE_{1 \rightarrow \tau}$.
After voxelization into spatiotemporal grids, denoted as $\cV_{0 \rightarrow \tau}$ and $\cV_{1 \rightarrow \tau}$, they are processed by the same event encoder $\enc_\cE$, as both of them describe motion from the source frame to the target interpolation time.
Similarly, the predicted pointmaps $(\pmap_0^0, \pmap_1^0)$ and the corresponding confidence $(\conf_0^0, \conf_1^0)$ are also encoded by the same type of small vision transformer (ViT) encoders ($\enc_\pmap$ and $\enc_\conf$) respectively.
The extracted features from the pointmap predictions as well as the event voxels are then gradually fused in the \interpr{} decoders $(\dec_{0 \rightarrow \tau}, \dec_{1 \rightarrow \tau})$ using zero convolution \citep{Zhang23adding}.

The decoders and heads of \interpr{} have the same structure as those of \alignr{} (but other choices are possible, as we show compatibility with \monstr{} in \cref{sec:experim:ablation}), and are initialized with \alignr{}'s weights.
Before being trained, \interpr{} exactly outputs $(\pmap_0^0, \pmap_1^0)$, i.e., the pointmaps of the two input frames.
As training proceeds, the information about the source pointmaps and about the motion from the source to the target time $\tau$ are gradually injected to the \interpr{} decoder.
In this way, we do not abruptly change \alignr{} decoders and heads, instead, the predicted pointmaps are progressively transported from the source to the target time, ensuring stable training.

In most EVFI methods \citep{Tulyakov21cvpr,Tulyakov22cvpr,Ma24timelenxl}, forward and backward events are simply voxelized into spatiotemporal volumes with identical channel dimensions and directly fed into the network for motion prediction.
The interpolation time $\tau$ is only implicitly encoded through temporal truncation of events, without explicit modeling.
As a result, the interpolation quality becomes highly sensitive to the interpolation time, especially in the presence of non-uniform motion or complex scene dynamics.
Inspired by \citet{Jin25cvpr}, we explicitly incorporate a time encoding into the prediction heads, conditioning the model on the target time $\tau \in (0,1)$.

\textbf{Training Objective.}
As shown in \cref{fig:pipeline}, we only train the \interpr{} part (i.e., the \alignr{} component is frozen), and use the confidence-aware scale-invariant 3D regression loss of DUSt3R.
We normalize the interpolated $\pmap_\tau$ and groundtruth pointmaps $\bar{\pmap}_\tau$ at $t=\tau$ using the scale factors calculated from the \alignr{}-predicted and groundtruth pointmaps at $t=0\, \text{and}\, 1$: $z = \text{norm}(\pmap_0, \pmap_1)$ and $\bar{z} = \text{norm}(\bar{\pmap}_0, \bar{\pmap}_1)$\footnote{where a bar denotes groundtruth and \emph{norm} computes the mean distance from the points to the world origin.}, respectively.
The final loss for image pair $(\img_0, \img_1)$ is:
\begin{equation}
\mathcal{L_\text{interp}} = 
\sum_{v \in \{0,1\}} 
\conf_{v \rightarrow \tau}
\Bigl\| \frac1{z} \pmap_{v \rightarrow \tau}
- \frac1{\bar{z}} \bar{\pmap}_{v \rightarrow \tau}
\Bigr\|
- \alpha \log \conf_{v \rightarrow \tau}
\end{equation}
where $\conf_{v \rightarrow \tau}$ is the pixel-wise confidence of $\pmap_{v \rightarrow \tau}$, and $\alpha$ is a hyper-parameter that controls the regularization term \citep{Wang24dus3r}.
This loss function is the sum of interpolation errors (3D weighted Euclidean distance) from both temporal directions.

\subsection{Coarse-to-Fine Global Alignment}
\label{sec:method:alignment}

After obtaining the pointmaps for all the frame and interpolation timestamps, we perform a coarse-to-fine global alignment to solve for the expected camera intrinsics, poses and depth at all those timestamps.
We illustrate the proposed coarse-to-fine alignment in the case of an image pair in \cref{fig:alignment}, where $\Theta_0$, $\Theta_\tau$ and $\Theta_1$ are the desired variables.
During coarse alignment, we align the source pointmaps produced by \alignr{} (in blue boxes) to solve for $\Theta_0$ and $\Theta_1$, which is the same as the global alignment in \alignr{}.
Note that dynamic scenes can be handled due to the capability of \alignr{}.
The objective function of the coarse alignment is \eqref{eq:monstr_loss}, i.e., the one in \monstr{}, where RAFT~\citep{Teed20eccv} is used to provide the optical flow required by the flow loss.

The fine alignment stage has two steps:
($i$) fixing the variables optimized in the coarse stage, aligning the interpolated pointmaps to the source pointmaps to solve for the variables at the target time, $\Theta_\tau$;
($ii$) finetuning all variables ($\Theta_0$, $\Theta_\tau$, $\Theta_1$) to further improve consistency.
In the second step, we add a scale-invariant coarse-translation prior term $\mathcal{L}_\text{trans}$ to \eqref{eq:monstr_loss} to preserve the shape of the coarse-frame trajectory while allowing one global scale degree of freedom, whose formulation is described in \eqref{eq:L_trans} in the supplementary.
Therefore, the final loss of the second step is $\mathcal{L}_\text{total} = \mathcal{L}_\text{\monstr{}} + w_\text{trans} \mathcal{L}_\text{trans}$.
In the fine alignment, the flow loss uses optical flow from E-RAFT~\citep{Gehrig21threedv}.
In practice, the target time $\tau$ can be set to any value between 0 and 1, and an arbitrary number of interpolated pointmaps can be fine-aligned, leading to continuous-time depth and pose estimation.

Following \dustr{}, we symmetrize every consecutive image pair $(\img_0, \img_1)$ and event sets $(\cE_{0 \rightarrow \tau}, \cE_{1 \rightarrow \tau})$ to obtain a new pair $(\img_1, \img_0)$ with $(\cE_{1 \rightarrow \tau}, \cE_{0 \rightarrow \tau})$.
We also use this new pair to interpolate the pointmaps $(\pmap_{1 \rightarrow \tau}^{1}, \pmap_{0 \rightarrow \tau}^1)$, where the target time is $1 - \tau$ and the output pointmaps are expressed in the camera coordinate of $\img_1$.
In this way, we obtain four pointmaps with confidence at the target position, which are inferred from two directions while expressed in two different camera coordinates.
These four measurements are then naturally fused according to pixel-wise confidence to obtain the estimation of depth and pose at $t=\tau$ in the fine alignment stage.

\section{Experiments}
\label{sec:experim}

\begin{table}[!t]
\centering
\caption{Depth estimation results.
\textbf{Best} and \underline{second-best} are highlighted.
\label{tab:depth}}
\setlength{\tabcolsep}{6pt}
\renewcommand{\arraystretch}{1.15}
\resizebox{\textwidth}{!}{%
\begin{tabular}{l | l | cc | cc | cc | cc}
\hline
\multirow{2}{*}{\textbf{Dataset}} & \multirow{2}{*}{\textbf{Method}}
& \multicolumn{2}{c|}{\textbf{skip = 1}}
& \multicolumn{2}{c|}{\textbf{skip = 3}}
& \multicolumn{2}{c|}{\textbf{skip = 7}}
& \multicolumn{2}{c}{\textbf{skip = 15}} \\
\cline{3-10}
&
& Abs Rel$\downarrow$ & $\delta < 1.25 \uparrow$
& Abs Rel$\downarrow$ & $\delta < 1.25 \uparrow$
& Abs Rel$\downarrow$ & $\delta < 1.25 \uparrow$
& Abs Rel$\downarrow$ & $\delta < 1.25 \uparrow$ \\
\Xhline{1.2pt}

\multirow{4}{*}{PointOdyssey}
& RIFE     & \num{0.097} & \unum{0.925} & \num{0.106} & \num{0.911} & \num{0.146} & \unum{0.891} & \num{0.196} & \unum{0.846} \\
& TimeLens & \unum{0.087} & \unum{0.925} & \unum{0.099} & \unum{0.912} & \num{0.142} & \num{0.858} & \num{0.215} & \num{0.777} \\
& VDM-EVFI & \num{0.108} & \num{0.889} & \num{0.113} & \num{0.875} & \unum{0.118} & \num{0.861} & \unum{0.137} & \num{0.839} \\
& \interpr{} (Ours)
& \bnum{0.0607} & \bnum{0.9541} & \bnum{0.0609} & \bnum{0.9572} & \bnum{0.0719} & \bnum{0.9434} & \bnum{0.1051} & \bnum{0.8863} \\
\Xhline{1.2pt}

\multirow{4}{*}{Sintel}
& RIFE     & \num{0.381} & \num{0.538} & \num{0.478} & \num{0.532} & \num{0.515} & \num{0.522} & \num{0.580} & \num{0.507} \\
& TimeLens & \num{0.377} & \unum{0.548} & \num{0.441} & \num{0.545} & \num{0.530} & \num{0.496} & \num{0.595} & \num{0.453} \\
& VDM-EVFI & \unum{0.366} & \num{0.542} & \unum{0.390} & \unum{0.553} & \unum{0.397} & \unum{0.559} & \unum{0.424} & \unum{0.543} \\
& \interpr{} (Ours)
& \bnum{0.3335} & \bnum{0.6101} & \bnum{0.3282} & \bnum{0.6212} & \bnum{0.3537} & \bnum{0.6125} & \bnum{0.4173} & \bnum{0.5726} \\
\Xhline{1.2pt}

\multirow{4}{*}{Bonn}
& RIFE     & \num{0.075} & \unum{0.956} & \num{0.079} & \unum{0.952} & \num{0.098} & \num{0.932} & \num{0.157} & \num{0.886} \\
& TimeLens & \num{0.077} & \unum{0.956} & \num{0.082} & \num{0.949} & \num{0.104} & \num{0.922} & \num{0.226} & \num{0.758} \\
& VDM-EVFI & \unum{0.074} & \bnum{0.959} & \unum{0.076} & \bnum{0.955} & \unum{0.076} & \unum{0.956} & \unum{0.076} & \bnum{0.956} \\
& \interpr{} (Ours)
& \bnum{0.0664} & \num{0.9493} & \bnum{0.0644} & \unum{0.9517} & \bnum{0.0641} & \bnum{0.9592} & \bnum{0.0715} & \unum{0.9554} \\
\Xhline{1.2pt}

\multirow{4}{*}{DSEC}
& RIFE     & \unum{0.1173} & \unum{0.8601} & \unum{0.1314} & \unum{0.8387} & \unum{0.1540} & \unum{0.8056} & \unum{0.1790} & \unum{0.7898} \\
& TimeLens & \num{0.1265} & \num{0.8371} & \num{0.1508} & \num{0.7889} & \num{0.2029} & \num{0.6944} & \num{0.3107} & \num{0.5083} \\
& VDM-EVFI & \num{0.1500} & \num{0.7861} & \num{0.1531} & \num{0.7801} & \num{0.1663} & \num{0.7443} & \num{0.1846} & \num{0.7050} \\
& \interpr{} (Ours)
& \bnum{0.0993} & \bnum{0.8967} & \bnum{0.0953} & \bnum{0.9104} & \bnum{0.1048} & \bnum{0.8830} & \bnum{0.1273} & \bnum{0.8580} \\
\hline
\end{tabular}}
\vspace{-2ex}
\end{table}

\subsection{Experimental Setup}
\label{sec:experim:setup}

\textbf{Implementation details.}
We train our model on six NVIDIA RTX A6000 GPUs with a total batch size of 12 for 50 epochs. 
We use the AdamW optimizer with a learning rate of $10^{-4}$. 
Training is performed solely on the PointOdyssey~\citep{Zheng23pointodyssey} dataset.
Each training sample consists of three images corresponding to timestamps $t=0$, $t=\tau$, and $t=1$ (i.e., a triplet). 
The temporal span of each triplet  ranges from 2 to 10 frames. 
We also symmetrize the triplets for training to encourage the consistency of the trained model, as \dustr{} does.
Before training, we precompute all valid triplets and construct a uniformly distributed sampling pool by balancing both the left-right temporal spans and the relative position of the middle frame.
This ensures that the model observes interpolation cases with different spans and relative positions evenly.
During training, each epoch randomly samples 20{,}000 triplets from this pool to train the model.

\textbf{Baselines.}
Similar to the evaluation of EVFI~\citep{Tulyakov21cvpr}, we skip $k = \{ 1, 3, 7, 15 \}$ frames, feed the remaining frames as well as the event data in between to \interpr{} to solve for depth and camera poses. 
These skips enable quantitative evaluation at groundtruth timestamps (see supplementary \cref{fig:continuous} for arbitrary-time qualitative results, e.g., $t=\{1/\pi,\sqrt{10}/4\}$).
Since our method is the first to address continuous-time depth and pose estimation tasks with events and frames, we create baselines based on concatenating state-of-the-art individual-task methods: 
we first leverage VFI methods: frame-based RIFE \citep{Huang22rife}, and event-based Timelens \citep{Tulyakov21cvpr} as well as VDM-EVFI \citep{Chen25cvpr} to interpolate skipped frames, and then input all frames to \alignr{} to solve for depth and poses.

\begin{table}[!t]
\centering
\caption{Pose evaluation results.
\textbf{Best} and \underline{second-best} are highlighted.
\label{tab:pose}}
\setlength{\tabcolsep}{6pt}
\renewcommand{\arraystretch}{1.15}
\resizebox{\textwidth}{!}{%
\begin{tabular}{l | l | ccc | ccc | ccc | ccc}
\hline
\multirow{2}{*}{\textbf{Dataset}} & \multirow{2}{*}{\textbf{Method}}
& \multicolumn{3}{c|}{\textbf{skip = 1}}
& \multicolumn{3}{c|}{\textbf{skip = 3}}
& \multicolumn{3}{c|}{\textbf{skip = 7}}
& \multicolumn{3}{c}{\textbf{skip = 15}} \\
\cline{3-14}
&
& ATE$\downarrow$ & RTE$\downarrow$ & RRE$\downarrow$
& ATE$\downarrow$ & RTE$\downarrow$ & RRE$\downarrow$
& ATE$\downarrow$ & RTE$\downarrow$ & RRE$\downarrow$
& ATE$\downarrow$ & RTE$\downarrow$ & RRE$\downarrow$ \\
\hline

\multirow{4}{*}{Sintel}
& RIFE
& \unum{0.301} & \unum{0.104} & \unum{1.259}
& \bnum{0.292} & \bnum{0.115} & \num{1.290}
& \unum{0.328} & \unum{0.146} & \num{1.675}
& \num{0.434} & \num{0.157} & \num{2.547} \\
& TimeLens
& \num{0.321} & \num{0.114} & \num{1.265}
& \num{0.385} & \num{0.174} & \num{1.412}
& \num{0.566} & \num{0.175} & \num{2.410}
& \num{0.599} & \unum{0.141} & \num{3.719} \\
& VDM-EVFI
& \num{0.319} & \num{0.127} & \num{1.314}
& \unum{0.339} & \unum{0.128} & \unum{1.193}
& \num{0.405} & \num{0.148} & \unum{1.351}
& \unum{0.379} & \bnum{0.118} & \bnum{1.496} \\
& \interpr{} (Ours)
& \bnum{0.2143} & \bnum{0.0885} & \bnum{0.8152}
& \bnum{0.2919} & \num{0.1401} & \bnum{0.9076}
& \bnum{0.2188} & \bnum{0.1014} & \bnum{1.2142}
& \bnum{0.2790} & \num{0.1538} & \unum{1.8689} \\
\hline

\multirow{4}{*}{Bonn}
& RIFE
& \unum{0.011} & \unum{0.007} & \unum{0.749}
& \unum{0.012} & \unum{0.007} & \unum{0.746}
& \unum{0.013} & \bnum{0.007} & \num{0.827}
& \unum{0.015} & \unum{0.008} & \unum{0.782} \\
& TimeLens
& \unum{0.011} & \num{0.008} & \num{0.762}
& \unum{0.012} & \num{0.008} & \num{0.765}
& \num{0.015} & \unum{0.009} & \unum{0.769}
& \num{0.017} & \num{0.010} & \num{1.482} \\
& VDM-EVFI
& \num{0.016} & \num{0.011} & \num{0.816}
& \num{0.017} & \num{0.011} & \num{0.792}
& \num{0.016} & \num{0.010} & \num{0.772}
& \num{0.019} & \num{0.010} & \num{0.835} \\
& \interpr{} (Ours)
& \bnum{0.0063} & \bnum{0.0061} & \bnum{0.5004}
& \bnum{0.0059} & \bnum{0.0066} & \bnum{0.5083}
& \bnum{0.0091} & \bnum{0.0068} & \bnum{0.5038}
& \bnum{0.0080} & \bnum{0.0063} & \bnum{0.5137} \\
\hline

\multirow{4}{*}{ECD}
& RIFE
& \unum{0.1749} & \unum{0.0651} & \unum{3.1437}
& \unum{0.1776} & \unum{0.0829} & \num{3.6705}
& \unum{0.1970} & \num{0.1042} & \num{3.6395}
& \num{0.2340} & \num{0.1044} & \num{4.0819} \\
& TimeLens
& \num{0.1837} & \unum{0.0651} & \num{3.1837}
& \num{0.2616} & \num{0.0981} & \num{3.6925}
& \num{0.3389} & \num{0.1045} & \num{3.6431}
& \num{0.3592} & \num{0.1126} & \num{3.7733} \\
& VDM-EVFI
& \num{0.2283} & \num{0.0842} & \num{3.7254}
& \num{0.2138} & \num{0.0836} & \unum{3.6441}
& \num{0.2259} & \unum{0.0925} & \unum{3.4931}
& \unum{0.2229} & \unum{0.0830} & \unum{3.3801} \\
& \interpr{} (Ours)
& \bnum{0.1212} & \bnum{0.0271} & \bnum{2.2130}
& \bnum{0.1379} & \bnum{0.0393} & \bnum{2.3462}
& \bnum{0.1723} & \bnum{0.0483} & \bnum{2.5118}
& \bnum{0.2178} & \bnum{0.0711} & \bnum{2.4784} \\
\hline

\multirow{4}{*}{EDS}
& RIFE
& \unum{0.0450} & \unum{0.0225} & \unum{0.4870}
& \unum{0.0460} & \unum{0.0227} & \num{0.5333}
& \bnum{0.0404} & \unum{0.0230} & \num{0.8935}
& \unum{0.0912} & \num{0.0529} & \num{1.6841} \\
& TimeLens
& \num{0.0467} & \num{0.0253} & \num{0.5286}
& \num{0.0513} & \num{0.0255} & \unum{0.5102}
& \unum{0.0700} & \num{0.0372} & \unum{0.6995}
& \num{0.2330} & \unum{0.0484} & \num{1.7757} \\
& VDM-EVFI
& \num{0.1031} & \num{0.0501} & \num{1.1635}
& \num{0.1036} & \num{0.0462} & \num{1.0580}
& \num{0.1115} & \num{0.0398} & \num{1.0162}
& \num{0.1063} & \bnum{0.0465} & \unum{1.0257} \\
& \interpr{} (Ours)
& \bnum{0.0242} & \bnum{0.0064} & \bnum{0.3456}
& \bnum{0.0226} & \bnum{0.0065} & \bnum{0.3386}
& \bnum{0.0403} & \bnum{0.0167} & \bnum{0.3464}
& \bnum{0.0855} & \num{0.0489} & \bnum{0.6593} \\
\hline
\end{tabular}}
\vspace{-2ex}
\end{table}

\subsection{Continuous-time Depth Estimation}
\label{sec:experim:depth}

\textbf{Evaluation datasets.}
We evaluate our approach on both synthetic and real-world benchmarks.
The real datasets include Bonn~\citep{Palazzolo19refusion} and DSEC~\citep{Gehrig21ral}, while the synthetic datasets include Sintel~\citep{Butler12eccv} and the validation split of PointOdyssey~\citep{Zheng23pointodyssey}.
Since Bonn, Sintel, and PointOdyssey do not contain event data, we generate events for them using vid2e~\citep{Gehrig20cvpr}.
For DSEC, we use recorded real events and spatially align RGB frames to them, then evaluate our method with real frame and event measurements.

\textbf{Evaluation metrics.}
Following prior works~\citep{Zhang25MonST3R,Lu25Align3R}, we first align the predicted depth maps to the groundtruth (GT) depth using a single scale and shift before computing the evaluation metrics.
To make the comparison of temporal consistency fair across different methods, the scale and shift are estimated once for the entire image sequence rather than independently for each frame.
We report two standard metrics: the absolute relative error (Abs Rel~$\downarrow$) and the fraction of inlier pixels with  $\delta < 1.25$ ($\uparrow$).
To highlight the performance on the interpolated frames, the metrics are calculated on the skipped frames.

\textbf{Comparison with baseline methods.}
\Cref{tab:depth} shows that \interpr{} achieves the lowest Abs Rel on all four datasets at every skip value, together with the highest inlier ratio on PointOdyssey, Sintel, and DSEC.
On PointOdyssey, it reduces Abs Rel over the strongest baseline by approximately 23\%--39\%, while on Bonn its accuracy remains nearly constant as the frame gap grows.

DSEC provides a particularly relevant test of generalization because it contains real images and event measurements from outdoor driving scenes, whereas training uses only synthetic PointOdyssey data.
\interpr{} leads both depth metrics across all temporal intervals, reducing Abs Rel by approximately 15\%--32\% relative to the strongest baseline.
The improvement remains substantial under sparse observations: at skips $3$, $7$, and $15$, the error reduction is approximately 27\%--32\%.
At skips $7$ and $15$, the inlier ratio also improves by approximately 9\%--10\% relative to the best baseline, indicating that the gains extend to a larger fraction of accurately reconstructed pixels.

As the frame gap increases, RIFE and TimeLens degrade markedly on DSEC, while \interpr{} retains lower error throughout.
This suggests that events provide useful motion cues for recovering intermediate geometry, especially when image observations are far apart.
Since TimeLens and VDM-EVFI also use events, the gains suggest that the benefit extends beyond simply adding the sensing modality.
\interpr{} couples event motion cues with source pointmaps during 3D prediction, allowing intermediate depth to be estimated directly within the geometric representation.

The results also show that these cues remain effective with real sensor data, supporting the transfer of our frame-event fusion from synthetic training to real-world continuous-time depth estimation.

\def\figWidth{0.145\linewidth}
\begin{figure}[t]
	\centering
    {\footnotesize
    \setlength{\tabcolsep}{0.5pt}
	\begin{tabular}{
	>{\centering\arraybackslash}m{0.3cm}
	>{\centering\arraybackslash}m{\figWidth}
	>{\centering\arraybackslash}m{\figWidth}
	>{\centering\arraybackslash}m{\figWidth}
	>{\centering\arraybackslash}m{\figWidth}
	>{\centering\arraybackslash}m{\figWidth}
	}

    \rotatebox{90}{\makecell{\scriptsize Frame}}
    &\imgbox{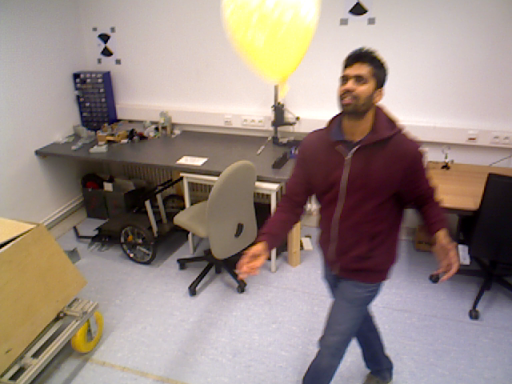}{(0.42,0.75) rectangle (0.64,0.99)}
    &\imgbox{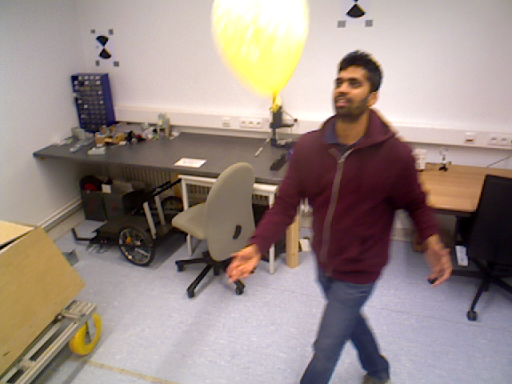}{(0.40,0.72) rectangle (0.62,0.99)}
    &\imgbox{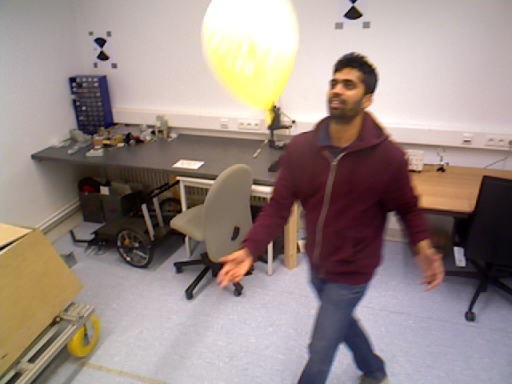}{(0.38,0.70) rectangle (0.60,0.99)}
    &\imgbox{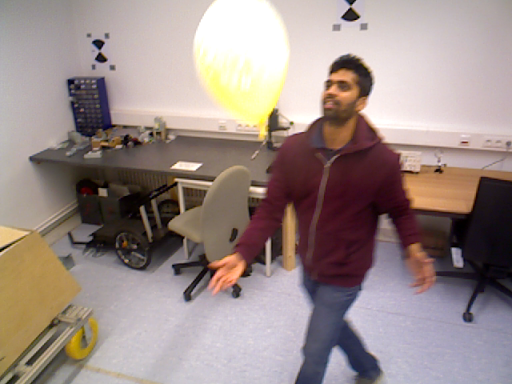}{(0.36,0.66) rectangle (0.58,0.99)}
    &\imgbox{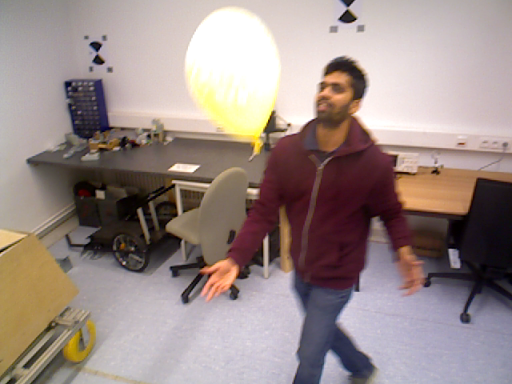}{(0.35,0.62) rectangle (0.56,0.99)}
	\\[-0.5ex]

    \rotatebox{90}{\makecell{\scriptsize RIFE}}
    &\imgbox{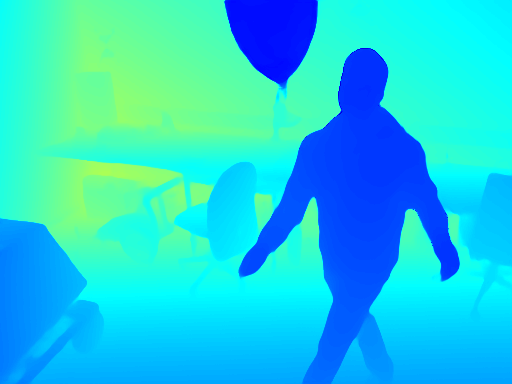}{(0.42,0.75) rectangle (0.64,0.99)}
    &\imgbox{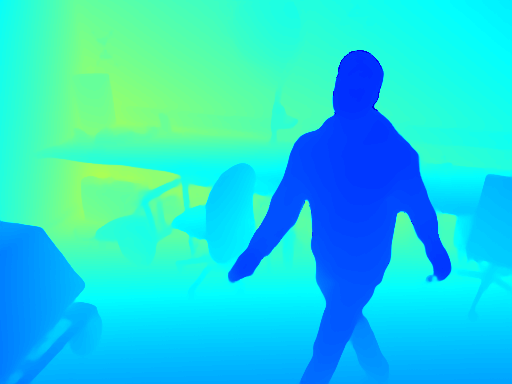}{(0.40,0.72) rectangle (0.62,0.99)}
    &\imgbox{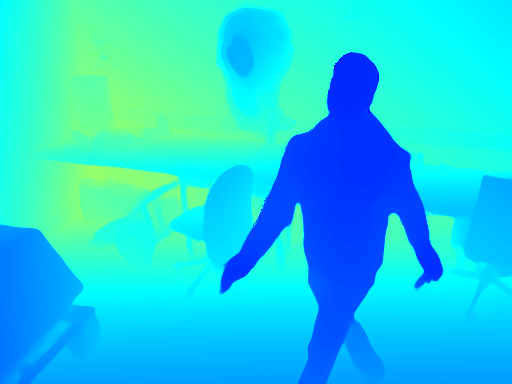}{(0.38,0.70) rectangle (0.60,0.99)}
    &\imgbox{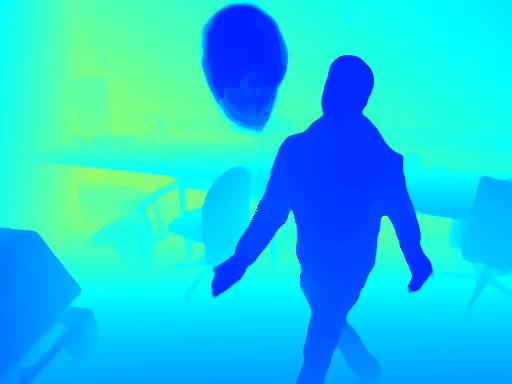}{(0.36,0.66) rectangle (0.58,0.99)}
    &\imgbox{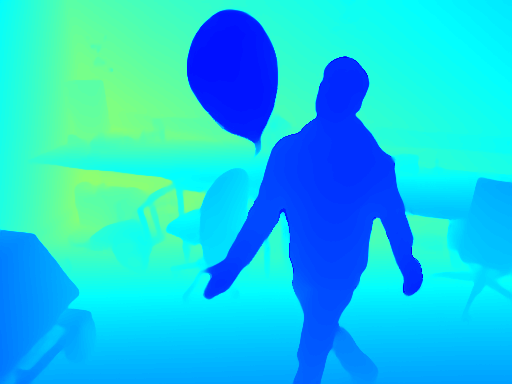}{(0.35,0.62) rectangle (0.56,0.99)}
	\\[-0.5ex]

    \rotatebox{90}{\makecell{\scriptsize TimeLens}}
    &\imgbox{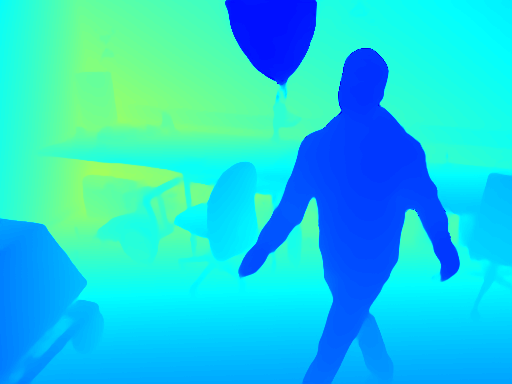}{(0.42,0.75) rectangle (0.64,0.99)}
    &\imgbox{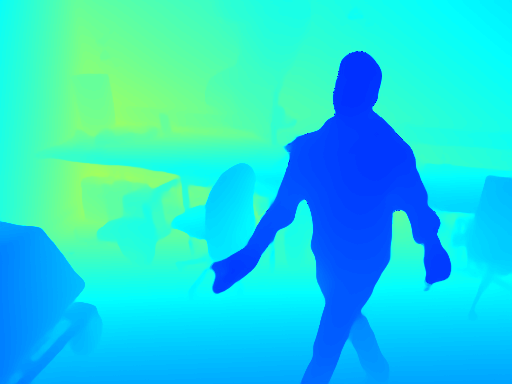}{(0.40,0.72) rectangle (0.62,0.99)}
    &\imgbox{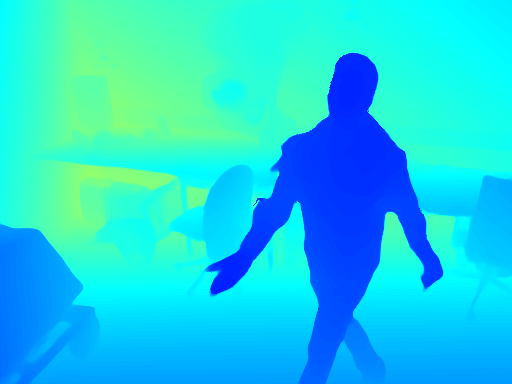}{(0.38,0.70) rectangle (0.60,0.99)}
    &\imgbox{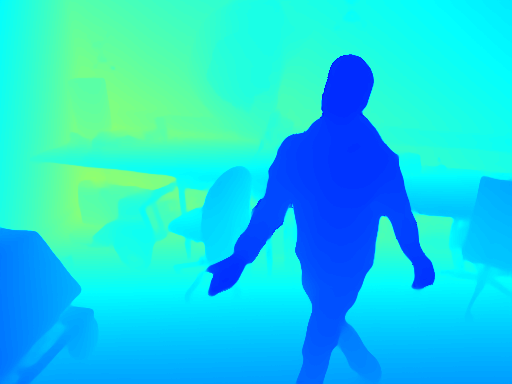}{(0.36,0.66) rectangle (0.58,0.99)}
    &\imgbox{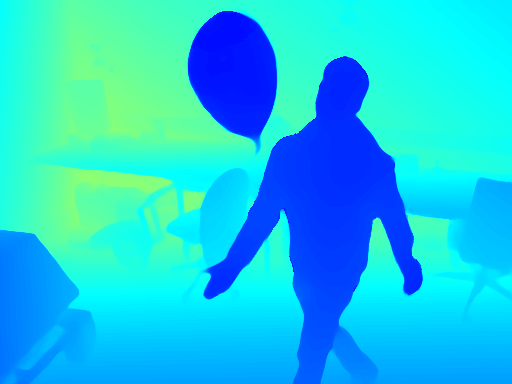}{(0.35,0.62) rectangle (0.56,0.99)}
	\\[-0.5ex]

    \rotatebox{90}{\makecell{\scriptsize VDM-EVFI}}
    &\imgbox{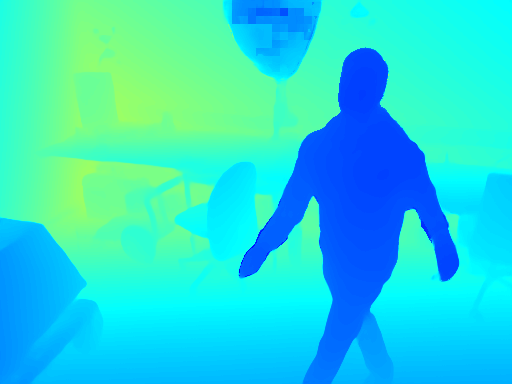}{(0.42,0.75) rectangle (0.64,0.99)}
    &\imgbox{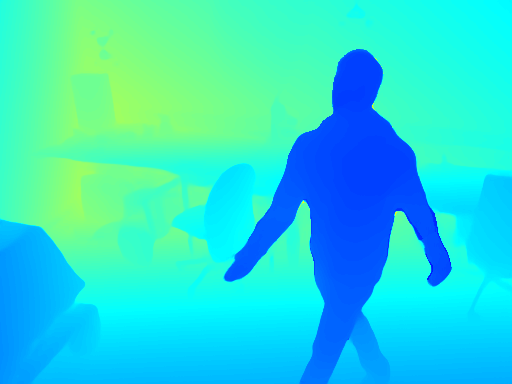}{(0.40,0.72) rectangle (0.62,0.99)}
    &\imgbox{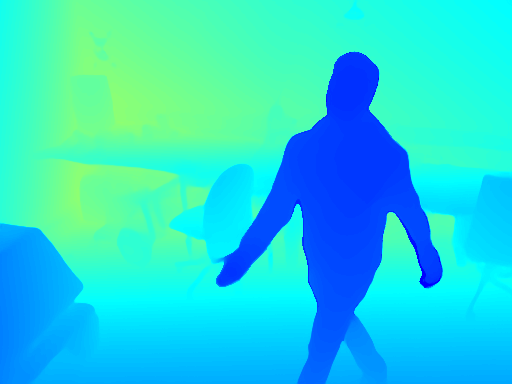}{(0.38,0.70) rectangle (0.60,0.99)}
    &\imgbox{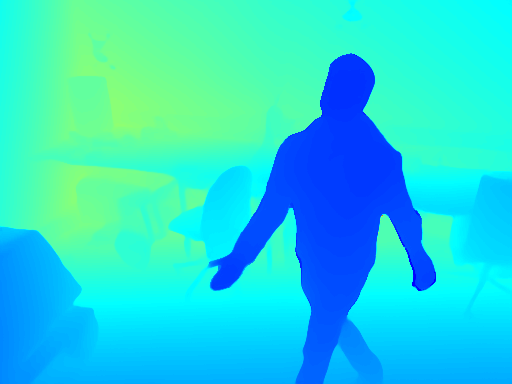}{(0.36,0.66) rectangle (0.58,0.99)}
    &\imgbox{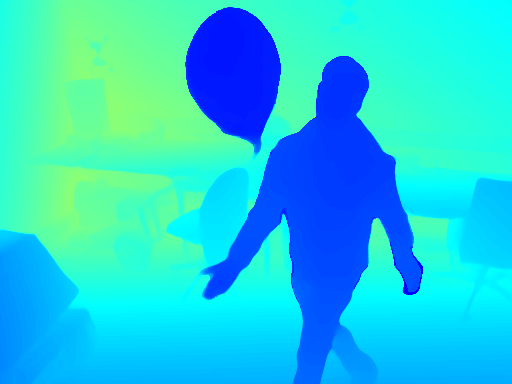}{(0.35,0.62) rectangle (0.56,0.99)}
	\\[-0.5ex]

    \rotatebox{90}{\makecell{\scriptsize Ours}}
    &\imgbox{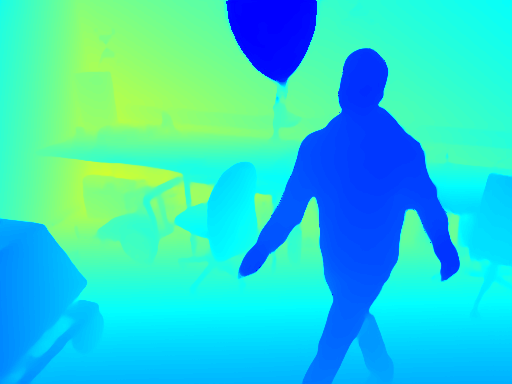}{(0.42,0.75) rectangle (0.64,0.99)}
    &\imgbox{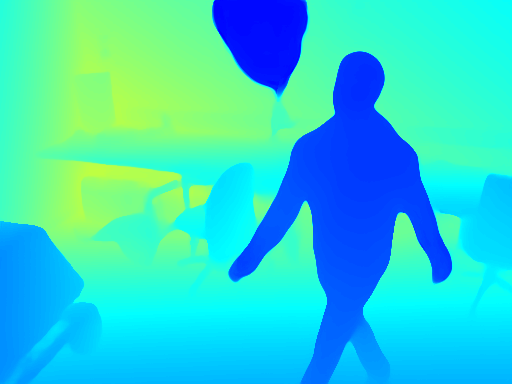}{(0.40,0.72) rectangle (0.62,0.99)}
    &\imgbox{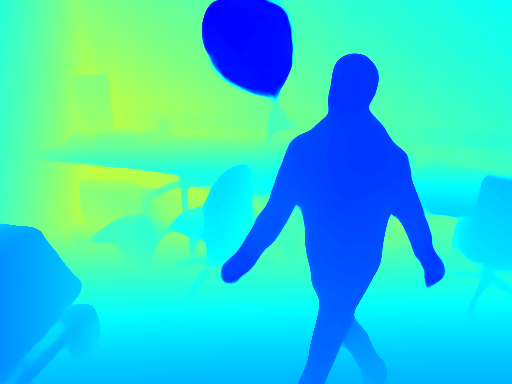}{(0.38,0.70) rectangle (0.60,0.99)}
    &\imgbox{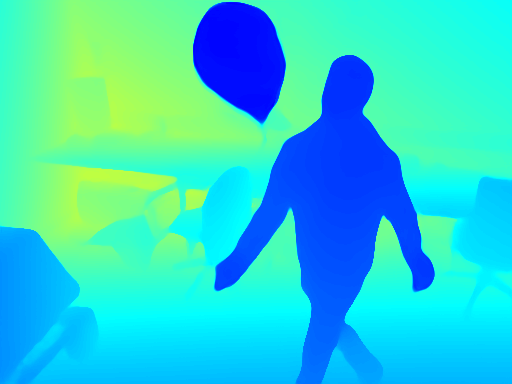}{(0.36,0.66) rectangle (0.58,0.99)}
    &\imgbox{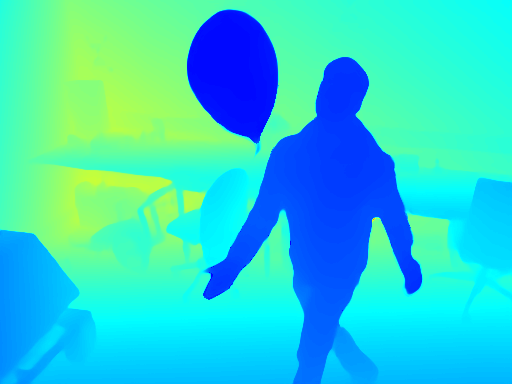}{(0.35,0.62) rectangle (0.56,0.99)}
	\\[-0.5ex]

        & \scriptsize $t = 0$
        & \scriptsize $t = 0.25$
        & \scriptsize $t = 0.5$
        & \scriptsize $t = 0.75$
        & \scriptsize $t = 1$
        \\[-0.5ex]

	\end{tabular}
	}
    \vspace{-1ex}
    \caption{Qualitative comparison of depth estimation on the Bonn dataset (skip = 3). \label{fig:bonn}}
    \vspace{-1ex}
\end{figure}

\def\figWidth{0.16\linewidth}
\begin{figure}[t]
	\centering
    {\footnotesize
    \setlength{\tabcolsep}{0.5pt}
	\begin{tabular}{
	>{\centering\arraybackslash}m{0.3cm}
	>{\centering\arraybackslash}m{\figWidth}
	>{\centering\arraybackslash}m{\figWidth}
	>{\centering\arraybackslash}m{\figWidth}
	>{\centering\arraybackslash}m{\figWidth}
	>{\centering\arraybackslash}m{\figWidth}
	}

    \rotatebox{90}{\makecell{\scriptsize Frames}}
	&\includegraphics[width=\linewidth]{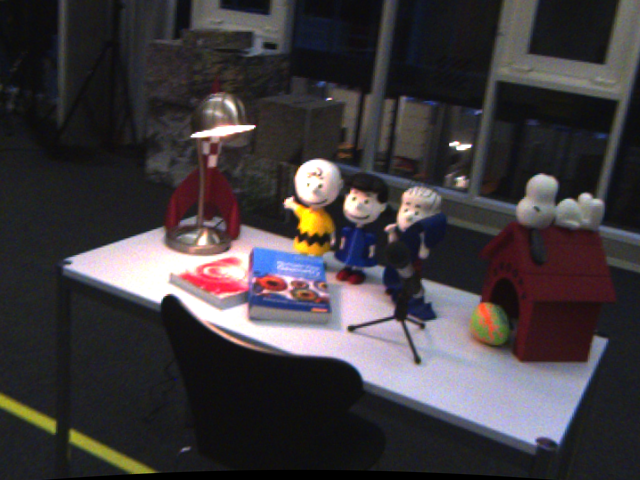}
	&\includegraphics[width=\linewidth]{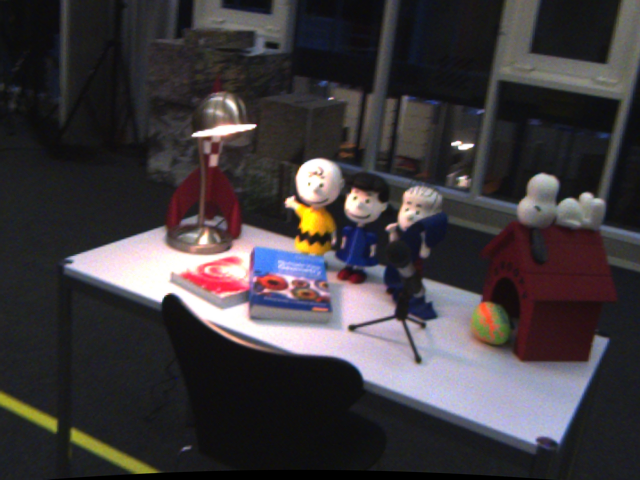}
	&\includegraphics[width=\linewidth]{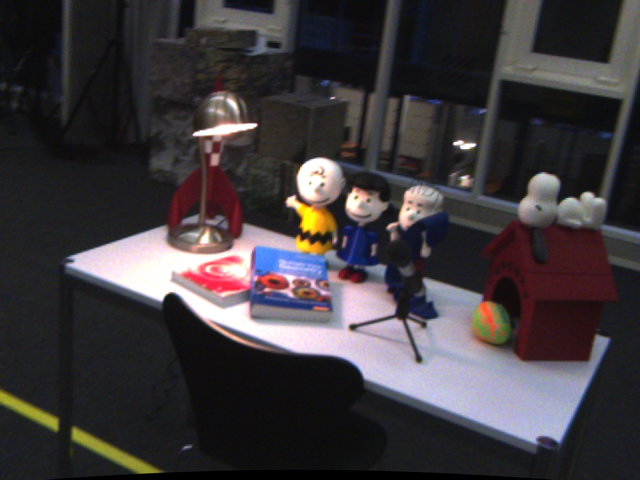}
	&\includegraphics[width=\linewidth]{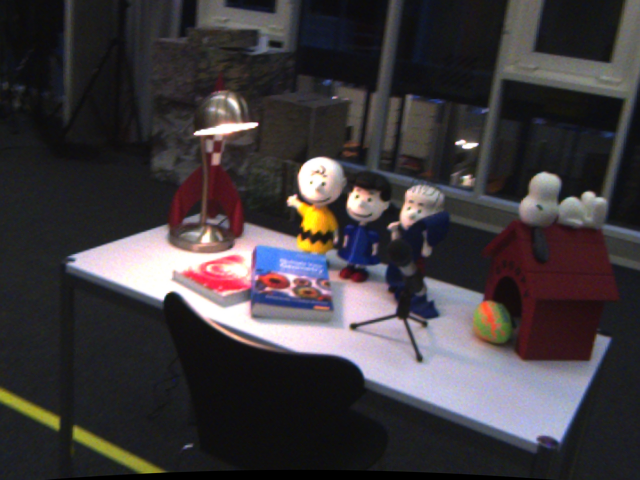}
	&\includegraphics[width=\linewidth]{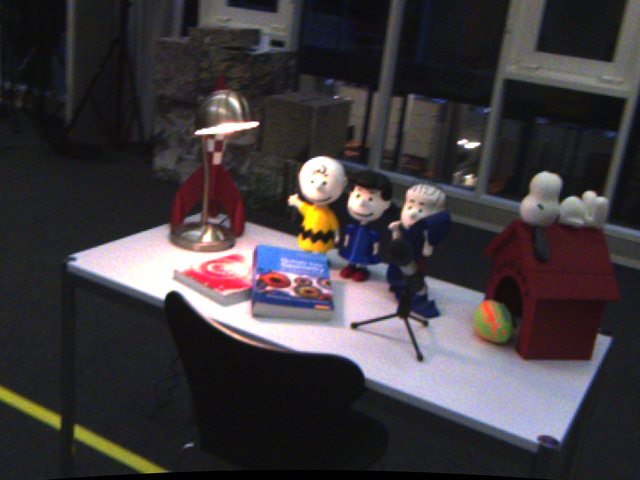}
	\\[-0.5ex]

    \rotatebox{90}{\makecell{\scriptsize Depth}}
	&\includegraphics[width=\linewidth]{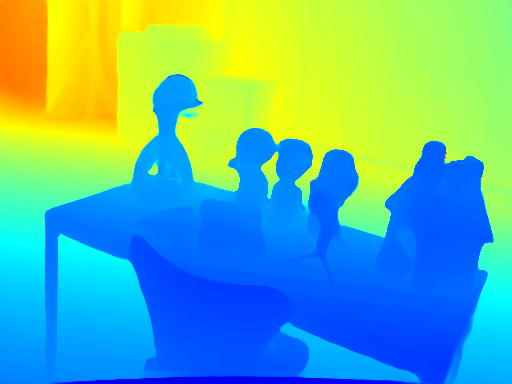}
	&\includegraphics[width=\linewidth]{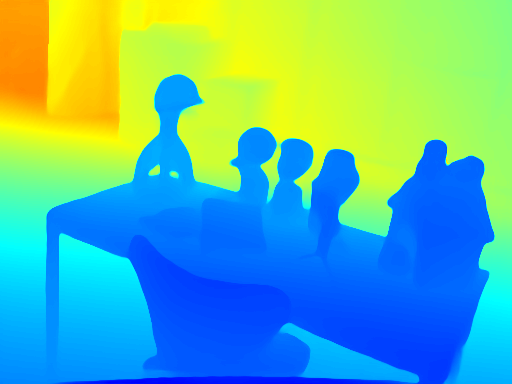}
	&\includegraphics[width=\linewidth]{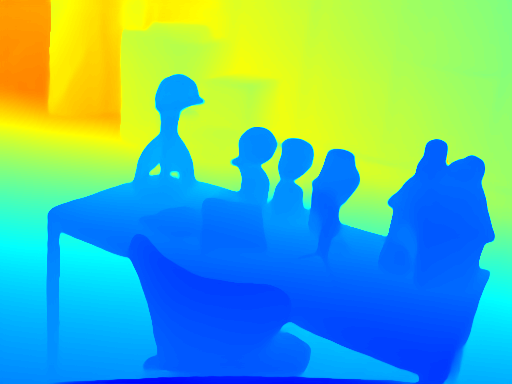}
	&\includegraphics[width=\linewidth]{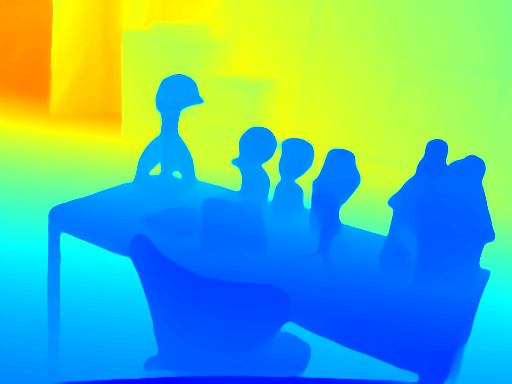}
	&\includegraphics[width=\linewidth]{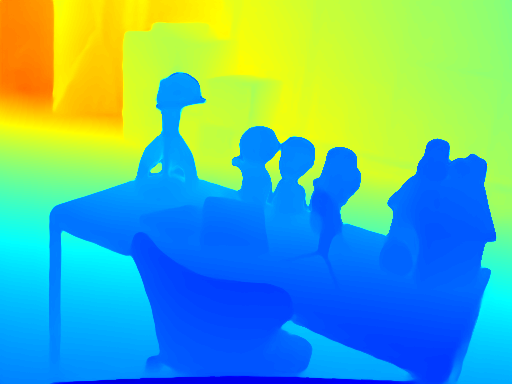}
	\\[-0.5ex]

        & \scriptsize $t = 0$
        & \scriptsize $t = 0.25$
        & \scriptsize $t = 0.5$
        & \scriptsize $t = 0.75$
        & \scriptsize $t = 1$
        \\[-0.5ex]

	\end{tabular}
	}
    \vspace{-1ex}
    \caption{Qualitative estimated depth results on the EDS dataset (skip = 3).\label{fig:eds}}
    \vspace{-1ex}
\end{figure}

\subsection{Continuous-time Pose Estimation}
\label{sec:experim:pose}

\textbf{Evaluation datasets.}
For camera pose evaluation, we largely follow the protocol used in Align3R and its preceding works.
The evaluation is conducted on four datasets: the real-world Bonn dataset, the synthetic Sintel dataset, and the real event camera datasets ECD~\citep{Mueggler17ijrr} and EDS~\citep{Hidalgo22cvpr}.
For ECD and EDS, we directly use the recorded event measurements, allowing us to evaluate continuous-time pose estimation with real event data.

\textbf{Evaluation metrics.}
Following prior studies~\citep{Zhang25MonST3R,Lu25Align3R}, we report three commonly used metrics:
ATE$\downarrow$ (Absolute Trajectory Error [m]), RTE$\downarrow$ (Relative Translation Error [m]), and RRE$\downarrow$ (Relative Rotation Error [$^\circ$]).
ATE measures the discrepancy between the estimated trajectory and the ground-truth trajectory.
RTE and RRE quantify the average local translation and rotation errors, respectively, computed between consecutive poses.

\begin{table}[!t]
\centering
\caption{\textbf{Ablation studies.}
Depth estimation on DSEC and camera pose estimation on EDS.
\label{tab:ablation}
}
\setlength{\tabcolsep}{0pt}
\newcolumntype{D}{>{\hspace{1pt}}c<{\hspace{1pt}}}
\newcolumntype{P}{>{\hspace{5pt}}c<{\hspace{5pt}}}
\renewcommand{\arraystretch}{1.15}
\resizebox{\textwidth}{!}{%
\begin{tabular}{>{\hspace{3.5pt}}l<{\hspace{3.5pt}} | DD | DD | DD || PPP | PPP | PPP}
\hline
\multirow{3}{*}{\textbf{Method}}
& \multicolumn{6}{c||}{\textbf{Depth estimation (DSEC)}}
& \multicolumn{9}{c}{\textbf{Camera pose estimation (EDS)}} \\
\cline{2-16}
& \multicolumn{2}{c|}{\textbf{skip = 1}}
& \multicolumn{2}{c|}{\textbf{skip = 3}}
& \multicolumn{2}{c||}{\textbf{skip = 7}}
& \multicolumn{3}{c|}{\textbf{skip = 1}}
& \multicolumn{3}{c|}{\textbf{skip = 3}}
& \multicolumn{3}{c}{\textbf{skip = 7}} \\
\cline{2-16}
& Abs Rel$\downarrow$ & $\delta\!<\!1.25 \uparrow$
& Abs Rel$\downarrow$ & $\delta\!<\!1.25 \uparrow$
& Abs Rel$\downarrow$ & $\delta\!<\!1.25 \uparrow$
& ATE$\downarrow$ & RTE$\downarrow$ & RRE$\downarrow$
& ATE$\downarrow$ & RTE$\downarrow$ & RRE$\downarrow$
& ATE$\downarrow$ & RTE$\downarrow$ & RRE$\downarrow$ \\
\hline

w/o events
& \num{0.1034} & \num{0.8902}
& \num{0.1077} & \num{0.8927}
& \num{0.1226} & \num{0.8555}
& \num{0.0284} & \num{0.0096} & \num{0.4543}
& \num{0.0332} & \num{0.0148} & \num{0.4411}
& \num{0.0897} & \num{0.0562} & \num{0.4661} \\

\monstr{} + \interpr{}
& \num{0.1217} & \num{0.8552}
& \num{0.1112} & \num{0.8803}
& \num{0.1180} & \bnum{0.8869}
& \bnum{0.0237} & \bnum{0.0053} & \bnum{0.3370}
& \num{0.0314} & \num{0.0078} & \num{0.3726}
& \num{0.0605} & \num{0.0331} & \num{0.4011} \\

\alignr{} + \interpr{}
& \bnum{0.0993} & \bnum{0.8967}
& \bnum{0.0953} & \bnum{0.9104}
& \bnum{0.1048} & \num{0.8830}
& \bnum{0.0242} & \num{0.0064} & \num{0.3456}
& \bnum{0.0226} & \bnum{0.0065} & \bnum{0.3386}
& \bnum{0.0403} & \bnum{0.0167} & \bnum{0.3464} \\
\hline
\end{tabular}}
\vspace{-3ex}
\end{table}

\def\pointmaprowheight{0.21670\linewidth}
\pointmapcomparison{3D reconstruction of a driving sequence fro m the DSEC dataset (skip = 3).}%
  {{\includegraphics[trim=176bp 67bp 176bp 48bp,clip,height=\pointmaprowheight]{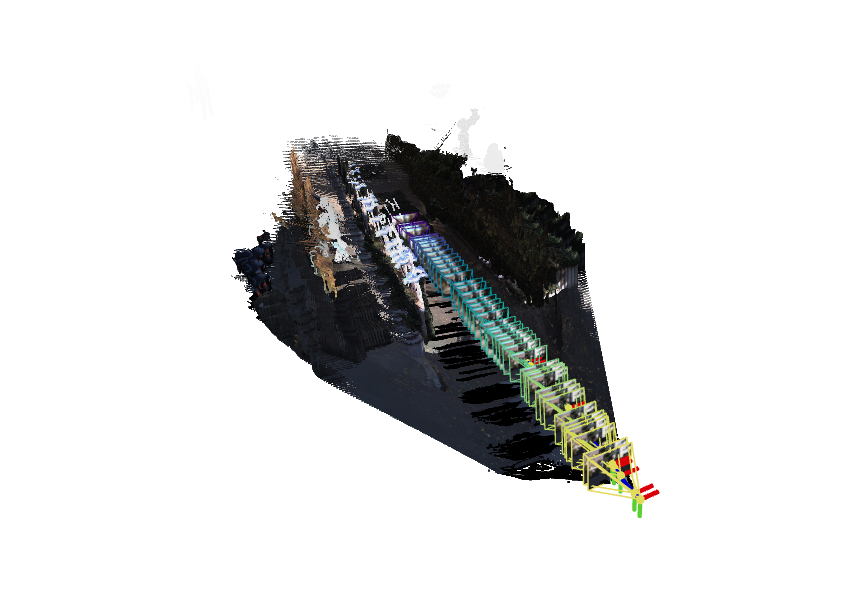}}}%
  {{\includegraphics[trim=224bp 83bp 133bp 53bp,clip,height=\pointmaprowheight]{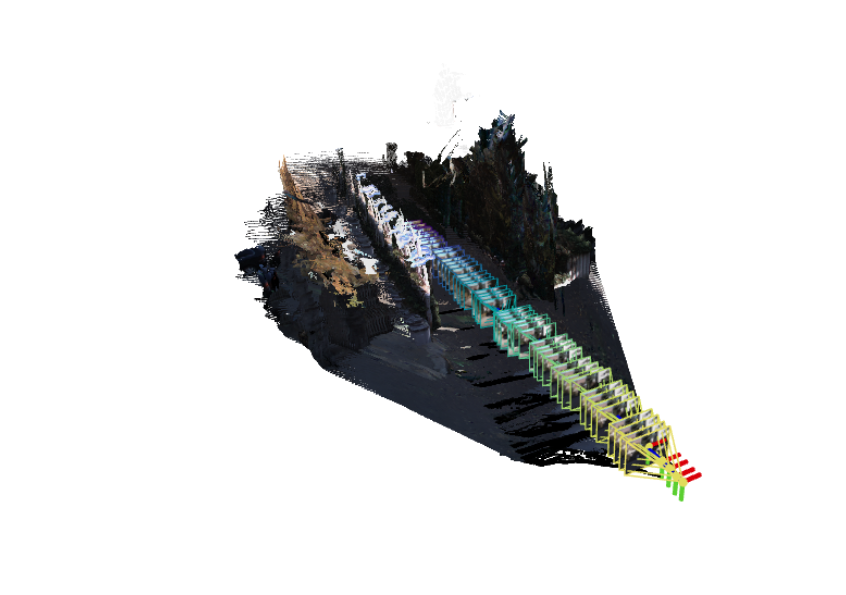}}}%
  {{\includegraphics[trim=150bp 122bp 158bp 99bp,clip,height=\pointmaprowheight]{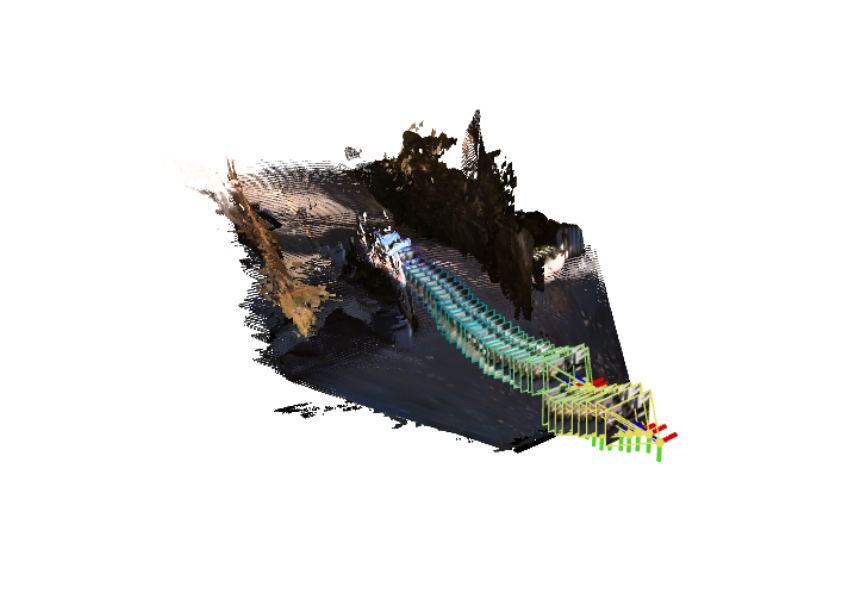}}}%
  {{\includegraphics[trim=181bp 50bp 239bp 50bp,clip,height=\pointmaprowheight]{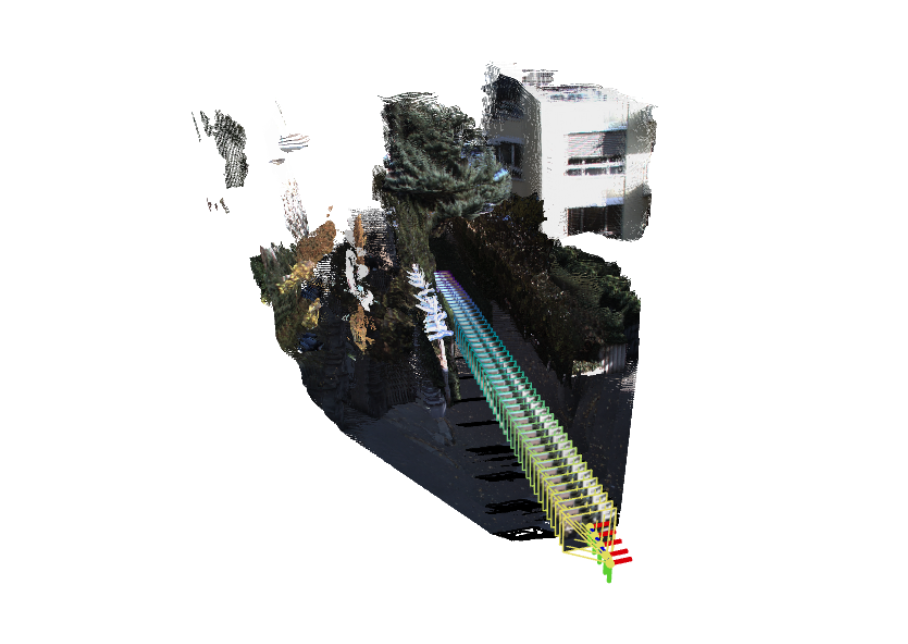}}}%
  {fig:pointmaps:dsec06}

\textbf{Comparison with baseline methods.}
\Cref{tab:pose} shows strong improvements in pose estimation across evaluated datasets.
\interpr{} reports the best ATE on all datasets at every skip, and best RTE and RRE in 13/16 and 15/16 settings, respectively.
The advantages transfer to real event camera data on ECD and EDS, despite training exclusively on synthetic sequences.

On ECD, \interpr{} achieves the best ATE, RTE, and RRE at every skip value.
Relative to the strongest baseline for each metric, it reduces RRE by 27\%--36\% and RTE by 14\%--58\% across the four skips.
The consistent improvements in both local motion and global trajectory accuracy indicate that event cues benefit the recovered camera poses as a whole.
At skip $=7$, the reductions in ATE, RTE, and RRE are 13\%, 48\%, and 28\%, respectively, showing that the advantage persists when the input frames are sparsely sampled.
On EDS, \interpr{} achieves the lowest ATE and RRE at every skip value, with particularly strong gains in rotation estimation: RRE decreases by 29\%--50\% relative to the strongest baseline, and the largest improvement occurs at skip $=7$.
At skips $1$ and $3$, ATE is reduced by 46\%--51\%, while RTE drops by about 71\%.
All these results show that the proposed method achieves precise pose estimation on real-world frames and events.

\vspace{-2ex}
\subsection{Qualitative Results}
\label{sec:experim:qualitative}
 
A qualitative comparison of estimated depth maps on Bonn is presented in \cref{fig:bonn}.
All baseline methods fail to estimate reasonable depth for the balloon.
More specifically, the balloon completely disappears in the depth maps of TimeLens and VDM-EVFI while partially shows up in those of RIFE.
For VDM-EVFI, the error is further propagated to the source frame ($t=0$).
This implies that the inaccuracy of the 2D VFI results is significantly amplified in the 3D geometry estimation, which leads to unreasonable and temporally inconsistent depth values.
In contrast, our method produces spatially smooth and temporally consistent depth estimation.

\Cref{fig:eds} shows qualitative results on the EDS dataset, estimated from real images and events.
The depth maps at $t=\{0.25, 0.5, 0.75\}$ recover the scene's fine geometric structures very well.
This shows our synthetically-trained model generalizes well to real images and events.
Finally, \Cref{fig:pointmaps:dsec06} depicts that our method retains a clearer building facade and a more coherent road surface than the baselines.
The camera poses (colored frusta) remain organized along the street, supporting joint reconstruction of scene geometry from real events.
More results are given in the Appendix.

\vspace{-1ex}
\subsection{Ablation Studies}
\label{sec:experim:ablation}

\Cref{tab:ablation} evaluates event input and backbone compatibility on DSEC depth and EDS pose estimation.

\textbf{Effects of Event Data.}
We remove the event input and encoder $\mathbf{E}_\mathcal{E}$ while keeping other training settings unchanged.
Events improve every reported depth and pose metric, with the Abs Rel reduction growing from 4\% to 15\% as the skip increases from 1 to 7, which shows event data becomes more important as the frame skip grows.
At skip $=7$, events reduce ATE and RTE by 55\% and 70\%, respectively, demonstrating their importance for recovering motion between sparse frames.

\textbf{Compatibility with \monstr{}.}
We replace \alignr{} with \monstr{} at inference time, retaining the \interpr{} model trained on \alignr{} without finetuning.
The resulting variant remains competitive across the evaluated skips and reduces RTE by 17\% at skip $=1$ relative to \alignr{} + \interpr{}.
This transfer demonstrates that \interpr{} can serve as a plug-and-play module for extending \dustr{}-like frame-based pointmap models to continuous-time estimation of camera motion and scene geometry.

\vspace{-3ex}
\section{Conclusion}
\vspace{-1ex}
\label{sec:conclusion}
This paper introduced \interpr{}, a novel framework that extends frame-based pointmap models to continuous-time 3D geometry estimation (depth maps and camera poses) by leveraging, for the first time, co-located asynchronous event camera data.
Thorough experiments showed that our approach consistently outperforms state-of-the-art baselines that rely on two-stage pipelines of video frame interpolation followed by frame-based geometry estimation.
We hope our work highlights the potential of combining event-based sensing with modern 3D visual foundation models and opens new directions for temporally continuous 3D perception in dynamic environments.

\clearpage
\section*{Acknowledgments}
This work has been partially supported by the German Federal Ministry of Research, Technology and Space (BMFTR) under the Robotics Institute Germany (RIG).
Funded by the Deutsche Forschungsgemeinschaft (DFG, German Research Foundation) under Germany’s Excellence Strategy – EXC 2002/1 ``Science of Intelligence'' – project number 390523135.

\appendix
\section*{Supplementary Material}

\section{Video}
The accompanying \textbf{video} shows more qualitative results of continuous-time 3D geometry estimation, which achieves higher frame-rate estimation of depth and camera poses than the original input videos (e.g. x2, x4 and x8).
Note that our method does not interpolate frames, so we use RIFE to interpolate frames for better visualization of the points.

\section{Arbitrary-Time Depth and Camera Pose Estimation}
\label{sec:supp:continuous}

The quantitative experiments in \cref{sec:experim} use skip values of 1, 3, 7 and 15 to evaluate predictions at timestamps with available ground truth.
These uniformly spaced evaluation times do not restrict the query times supported by \interpr{}.
Given two source frames at normalized times $t=0$ and $t=1$ and the events between them, our model accepts an arbitrary query time $t\in(0,1)$, predicts the corresponding pointmaps, and recovers depth and camera pose through global alignment (\cref{sec:method:alignment}).

\begin{figure}[h]
    \centering
    \begin{minipage}[t]{0.195\linewidth}
        \centering
        \includegraphics[width=\linewidth]{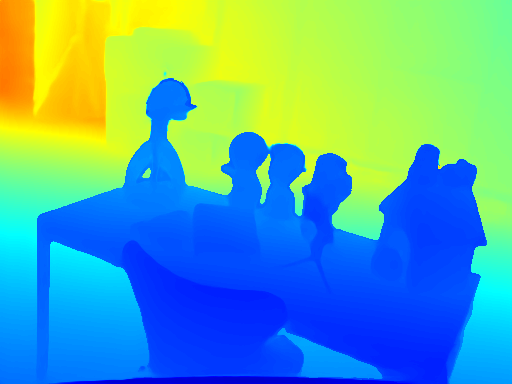}\\
        {\small $t=0$}
    \end{minipage}\hfill
    \begin{minipage}[t]{0.195\linewidth}
        \centering
        \includegraphics[width=\linewidth]{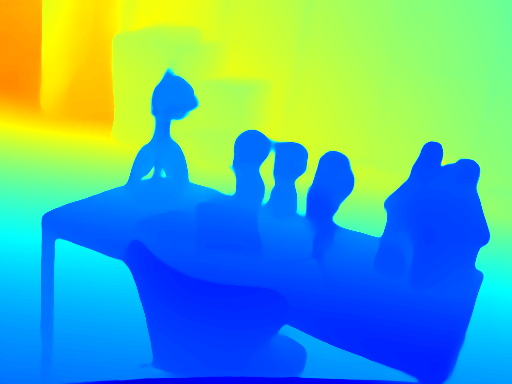}\\
        {\small $t=1/\pi$}
    \end{minipage}\hfill
    \begin{minipage}[t]{0.195\linewidth}
        \centering
        \includegraphics[width=\linewidth]{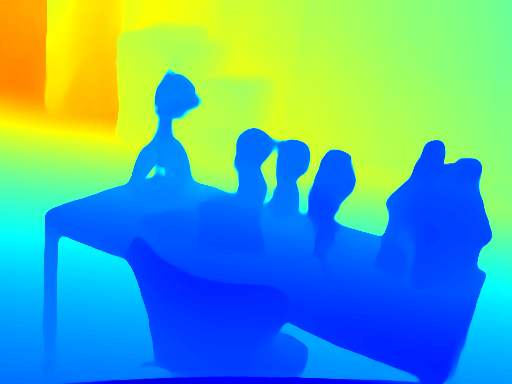}\\
        {\small $t=0.5$}
    \end{minipage}\hfill
    \begin{minipage}[t]{0.195\linewidth}
        \centering
        \includegraphics[width=\linewidth]{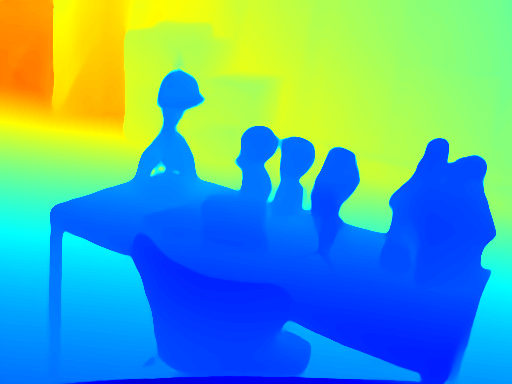}\\
        {\small $t=\sqrt{10}/4$}
    \end{minipage}\hfill
    \begin{minipage}[t]{0.195\linewidth}
        \centering
        \includegraphics[width=\linewidth]{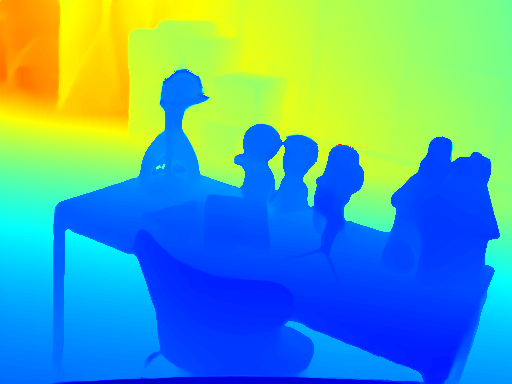}\\
        {\small $t=1$}
    \end{minipage}
    \caption{Continuous-time depth results with \interpr{}.
    The depth maps at $t=0$ and $t=1$ are shown together with predictions at $t=1/\pi$, $t=0.5$ and $t=\sqrt{10}/4$, illustrating inference beyond a fixed, uniformly spaced temporal grid.}
    \label{fig:continuous}
\end{figure}

\Cref{fig:continuous} illustrates this capability through depth predictions at $t=1/\pi$, $t=0.5$ and $t=\sqrt{10}/4$, alongside the depth maps at the two source timestamps.
In particular, $1/\pi$ and $\sqrt{10}/4$ lie outside the uniformly spaced grids used in our quantitative evaluation.
These results demonstrate that \interpr{} can estimate 3D geometry at user-specified intermediate times, beyond the fixed frame intervals used for benchmarking.

\section{Related Work}
\label{sec:supp:related}

\Cref{tab:related} provides a detailed comparison of state-of-the-art 3D geometry and video depth models, with frame-based, event-based or combined input.
The columns indicate whether the methods are used to interpolate video depth, camera poses (i.e., ego-motion), can work on dynamic scenes and interpolate at arbitrary times (last column).
The table reveals that our method is the first pointmap-based framework for arbitrary-time depth and camera pose estimation from frames
and events.

\begin{table}[t]
\centering
\caption{
\emph{Comparison of the main characteristics of Frame-based and Event-aided 3D Geometry / Video Depth Methods (i.e., deep learning models).}
\label{tab:related}
\vspace{0.5ex}
}
\begin{adjustbox}{max width=\linewidth}
\setlength{\tabcolsep}{3pt}
\renewcommand{\arraystretch}{1.2}
\begin{threeparttable}
\begin{tabular}{lccccccc}
\toprule
\textbf{Method} & \textbf{Year} & \textbf{Type} & \textbf{Input} & \textbf{Video} & \textbf{Pose} & \textbf{Dynamic Scene} & \textbf{Arbitrary-time} \\
\midrule
VideoDepthAnything \citep{Chen25VideoDepthAnything} & 2025 & FF & F & \yesmark & \nomark & \yesmark & \nomark \\ 
DepthAnything-v3 \citep{Lin25depthanything3} & 2025 & FF & F & \yesmark & \yesmark & \yesmark & \nomark \\ 
ChronoDepth \citep{Shao25cvpr} & 2025 & Diff & F & \yesmark & \nomark & \yesmark & \nomark \\ 
DepthCrafter \citep{Hu25cvpr} & 2025 & Diff & F & \yesmark & \nomark & \yesmark & \nomark \\ 
\dustr{} \citep{Wang24dus3r} & 2024 & PM & F & \yesmark & \yesmark & \nomark & \nomark \\ 
\monstr{} \citep{Zhang25MonST3R} & 2025 & PM & F & \yesmark & \yesmark & \yesmark & \nomark \\ 
\alignr{} \citep{Lu25Align3R} & 2025 & PM & F & \yesmark & \yesmark & \yesmark & \nomark \\
DynaDUSt3R~\citep{Jin25cvpr} & 2025 & PM & F & \yesmark & \nomark & \yesmark & \nomark \\ 

EventVGGT \citep{Ren26eventvggt} & 2026 & FF & E & \yesmark & \yesmark & \yesmark & \nomark \\
Event3R \citep{Huang26event3r} & 2026 & PM & E & \nomark & \yesmark & \nomark & \nomark \\ 
\eagr{} \citep{Wu25eag3r} & 2025 & PM & E+F & \yesmark & \yesmark & \yesmark & \nomark \\ 
\interpr{} (Ours) & 2026 & PM & E+F & \yesmark & \yesmark & \yesmark & \yesmark \\ 
\bottomrule
\end{tabular}
\begin{tablenotes}[para]
\item[1] FF $\equiv$ Feed-forward ViT;  Diff $\equiv$ Diffusion; PM $\equiv$ Pointmap-based Model; F $\equiv$ Frames; E $\equiv$ Events.
\end{tablenotes}
\end{threeparttable}
\end{adjustbox}
\end{table}

\section{Computational Complexity and Runtime Evaluation}
\label{sec:supp:effort}
\interpr{} has 370.98 million parameters.
\Cref{tab:runtime_combined} compares inference and alignment times for different numbers of interpolations, measured on a single NVIDIA A6000 GPU with 48\,GB of memory.

\begin{table}[!t]
\centering
\caption{\textbf{Computational effort.}
Inference and alignment times in seconds for $M$ interpolations.
\label{tab:runtime_combined}}
\small
\setlength{\tabcolsep}{8pt}
\renewcommand{\arraystretch}{1.15}
\begin{tabular}{l|l|cccc}
\hline
\multirow{2}{*}{Method} & \multirow{2}{*}{Stage} & \multicolumn{4}{c}{Number of interpolations $M$} \\
\cline{3-6}
& & 1 & 3 & 5 & 7 \\
\hline
\multirow{2}{*}{\interpr{}} 
    & Inference      & 0.285 & 0.426 & 0.635 & 0.725 \\
    & Alignment & 7.747 & 10.342 & 12.331 & 14.973 \\
\hline
\multirow{2}{*}{Baseline} 
    & Inference       & 0.360 & 1.030 & 1.999 & 3.287 \\
    & Alignment & 12.878 & 14.126 & 16.553 & 20.511 \\
\hline
\end{tabular}
\end{table}

Like other pointmap-based methods such as \dustr{}, \monstr{} and \alignr{}, \interpr{} requires global alignment and operates offline rather than in real time.
Its measured runtime grows approximately linearly with the number $M$ of interpolation times.
As illustrated in \cref{fig:alignment}, each interpolated timestamp is paired only with the two source frames, adding four pairwise pointmap constraints (four edges) to the fine-alignment graph.
The number of added constraints therefore scales as $O(M)$; for fixed image resolution and optimization iterations, their computational cost grows linearly with $M$.
The two-stage baseline method processes more frames and requires more inference calls and a larger alignment graph to produce the same number of intermediate estimates.
Consequently, \interpr{} is faster in both stages across all tested settings; with seven interpolations, inference is approximately $4.5\times$ faster and alignment takes 27\% less time than the baseline method.

\section{Limitations}
\label{sec:limitations}
Despite its effectiveness, our approach has limitations. 
First, \interpr{} performs pointmap interpolation directly on the outputs produced by the underlying frame-based model. 
When the source pointmaps are unreliable, e.g., due to motion blur or challenging illumination conditions in the input frames, the interpolation quality may also degrade. 
This issue could be mitigated by leveraging
event data to enhance the input images before geometry estimation, e.g., through event-guided deblurring or image enhancement, which could improve the quality of the predicted pointmaps and consequently benefit the interpolation process.

Second, the proposed method relies on event streams to capture motion cues between frames. 
However, when the scene contains flickering or rapidly changing light sources, the resulting events may not correspond to true scene motion. 
In such cases, the event signals can introduce artifacts and affect the interpolation results. 
This limitation could be addressed by detecting and removing such non-motion--caused events, or by injecting additional knowledge (e.g., priors).

\section{Detailed Description of Datasets and Evaluation Protocol}
\label{sec:supp:datasets}

\textbf{Depth estimation.}
The Bonn dataset is a real-world RGB-D SLAM benchmark consisting of 24 dynamic sequences involving human activities such as moving boxes or interacting with balloons.
Following the evaluation protocol of previous works~\citep{Zhang25MonST3R,Lu25Align3R}, we test our model on five sequences with an average length of 110 frames.
Sintel is a synthetic dynamic dataset composed of 23 sequences with roughly 50 frames each.
The PointOdyssey validation set contains 15 dynamic scenes with multiple moving foreground objects, and the first 110 frames of each sequence are evaluated.
We additionally evaluate on DSEC to assess the performance of continuous-time depth estimation in real outdoor scenes with real event measurements.
We select eight 50-frame clips from distinct sequences in the DSEC training split (so that ground truth depth is available for evaluations), covering urban and rural driving under varied illumination and with visible scene motion.
Event camera disparity groundtruth is available at half the 20 Hz RGB frame rate, so we retain every second left RGB frame, yielding 10 Hz clips in which each frame has a corresponding disparity map.
We warp the rectified left RGB frames into the rectified left-event camera grid using a calibration-derived, rotation-only homography.
We convert the provided event camera disparity to metric camera-$Z$ depth using the supplied calibration; invalid disparity yields zero depth, and evaluation includes the pixels with depth in $(10^{-3},70)$ m.

\textbf{Pose estimation.}
For the Bonn dataset, following prior works, we evaluate on five sequences and select 30 frames from each sequence for pose estimation.
For the Sintel dataset, we adopt the same filtering strategy as previous methods and exclude sequences that are static or only exhibit simple linear motion, resulting in 14 test sequences.
We additionally evaluate on ECD and EDS to examine whether the proposed approach generalizes from simulated events to real event camera measurements.
For ECD, we select six 100-frame clips, two each from \texttt{boxes\_6dof}, \texttt{dynamic\_6dof}, and \texttt{poster\_6dof}.
For EDS, we select eight 100-frame clips, two each from \texttt{01\_peanuts\_light}, \texttt{02\_rocket\_earth\_light}, \texttt{08\_peanuts\_running}, and \texttt{14\_ziggy\_in\_the\_arena}.

\section{Additional Implementation Details}
\label{sec:supp:implementation}

\textbf{ViT encoders for pointmaps, confidence and events.}
For each input modality, including the pointmap $\pmap$, confidence $\conf$, and event voxel representation $\mathcal{V}$, we first apply a standard patch embedding operation to partition the input into a sequence of patches:
\begin{equation}
\mathbf{W}' = \mathrm{PatchEmbed}(\mathbf{W}),
\end{equation}
where $\mathbf{W}' \in \mathbb{R}^{H' \times W' \times C}$ denotes the patch-embedded representation of the input modality $\mathbf{W}$. 
The resulting patch tokens are then processed by a Vision Transformer encoder to extract hierarchical feature representations $\hat{F}^{(1)}, \hat{F}^{(2)}, \ldots, \hat{F}^{(s)}$.
These multi-scale features are subsequently injected into the \interpr{} decoder to facilitate feature aggregation and refinement.

\textbf{Feature fusion using zero convolution.}
To preserve the prediction capability of the original pointmap model, we adopt zero convolution layers~\citep{Zhang23adding} to fuse the extracted features within the \interpr{} decoder:
\begin{equation}
\hat{E}^{(l)} = \mathrm{ZeroConv}(\hat{F}^{(l)}) + E^{(l)}, \quad l = 1,2,\ldots,s,
\end{equation}
where $E^{(l)}$ denotes the feature map produced by the \alignr{} decoder at the $l$-th layer, while $\hat{E}^{(l)}$ represents the fused feature map after incorporating the features derived from the input modalities (pointmaps, confidence maps, and events).

\def\figWidth{0.19\linewidth}
\begin{figure}[t]
	\centering
    {\footnotesize
    \setlength{\tabcolsep}{0.5pt}
	\begin{tabular}{
	>{\centering\arraybackslash}m{0.3cm}
	>{\centering\arraybackslash}m{\figWidth}
	>{\centering\arraybackslash}m{\figWidth}
	>{\centering\arraybackslash}m{\figWidth}
	>{\centering\arraybackslash}m{\figWidth}
	>{\centering\arraybackslash}m{\figWidth}
	}

    \rotatebox{90}{\makecell{\scriptsize Frames}}
	&\includegraphics[width=\linewidth]{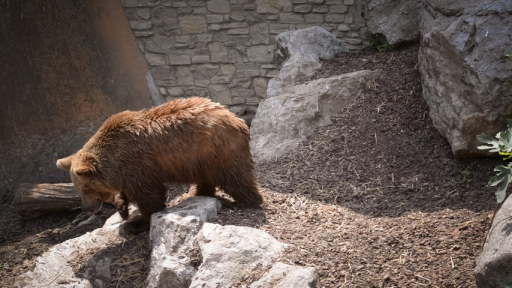}
	&\includegraphics[width=\linewidth]{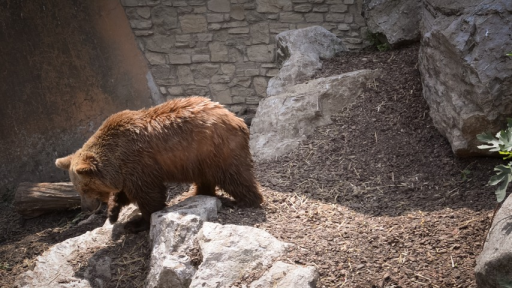}
	&\includegraphics[width=\linewidth]{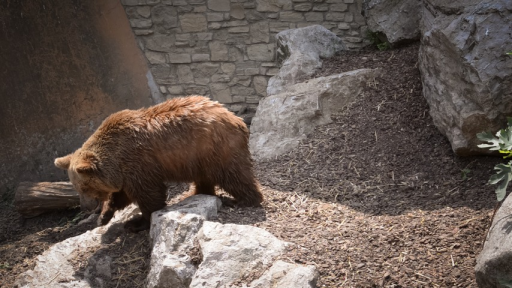}
	&\includegraphics[width=\linewidth]{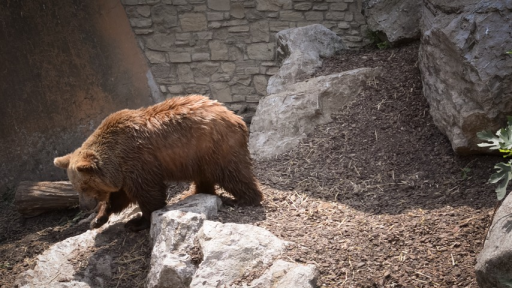}
	&\includegraphics[width=\linewidth]{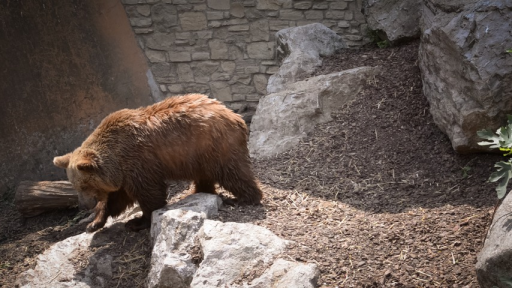}
	\\[-0.5ex]

    \rotatebox{90}{\makecell{\scriptsize Depth}}
	&\includegraphics[width=\linewidth]{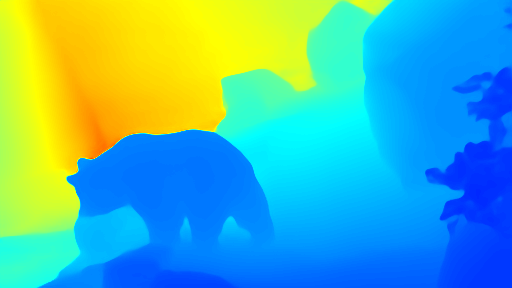}
	&\includegraphics[width=\linewidth]{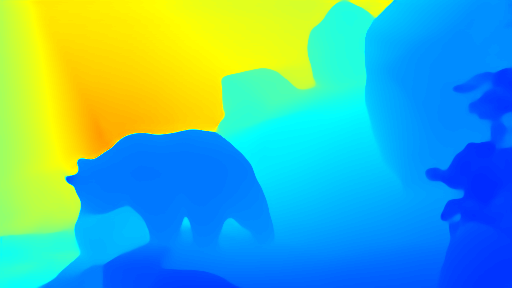}
	&\includegraphics[width=\linewidth]{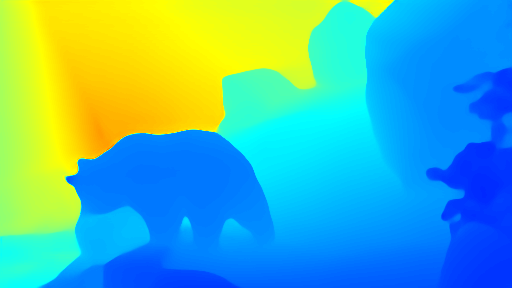}
	&\includegraphics[width=\linewidth]{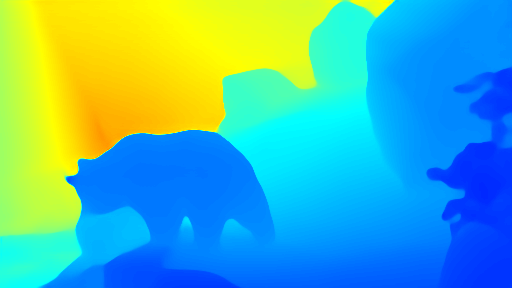}
	&\includegraphics[width=\linewidth]{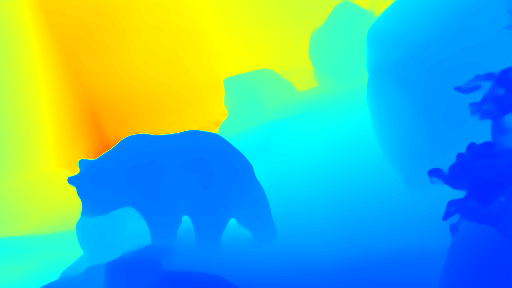}
	\\[-0.5ex]

        & \scriptsize $t = 0$
        & \scriptsize $t = 0.25$
        & \scriptsize $t = 0.5$
        & \scriptsize $t = 0.75$
        & \scriptsize $t = 1$
        \\[-0.5ex]

	\end{tabular}
	}
    \caption{Qualitative results of depth estimation on the DAVIS \texttt{bear} sequence (skip = 3). \label{fig:davis}}
\end{figure}

\textbf{Coarse-to-fine global alignment.}
In both coarse and fine stages of the proposed global alignment, we optimize the parameters for 300 iterations using the Adam optimizer. 
The learning rate is initialized at 0.01 and follows a linear decay schedule throughout the optimization process.
The loss weights are set as follows: $w_\text{smooth}=0.01, w_\text{flow}=0.01$.

\textbf{Scale-invariant coarse-translation prior.}
The first step of fine alignment keeps the coarse-frame geometry fixed while optimizing the interpolated frames. 
In the second step, jointly releasing the coarse and interpolated geometry improves global consistency, but can also cause the coarse trajectory to drift. 
We therefore use the first-step coarse trajectory as a soft reference for its local translational structure, rather than fixing its poses again. Let $\mathbf{c}_k^{(1)}$ and $\mathbf{c}_k^{(2)}$ denote the camera centers of the $k$-th coarse frame after the first and during the second step, respectively, and define the adjacent displacements $\mathbf{d}_k^{(j)}=\mathbf{c}_{k+1}^{(j)}-\mathbf{c}_k^{(j)}$ for $j\in\{1,2\}$. 
We fit a single positive scale to all $K-1$ coarse-frame displacements,
\begin{equation}
s^{\star}=\max\!\left(\epsilon,\frac{\sum_{k=1}^{K-1}(\mathbf{d}_k^{(2)})^{\top}\mathbf{d}_k^{(1)}}{\max\!\left(\sum_{k=1}^{K-1}\|\mathbf{d}_k^{(2)}\|_2^2,\epsilon\right)}\right),
\end{equation}
and penalize the scale-aligned residuals:
\begin{equation}
\mathcal{L}_{\mathrm{trans}}=
\frac{\frac{1}{K-1}\sum_{k=1}^{K-1}\left\|s^{\star}\mathbf{d}_k^{(2)}-\mathbf{d}_k^{(1)}\right\|_2}
{\max\!\left(\sqrt{\frac{1}{K-1}\sum_{k=1}^{K-1}\|\mathbf{d}_k^{(1)}\|_2^2},\epsilon\right)}.
\label{eq:L_trans}
\end{equation}
Here $\epsilon$ denotes a small numerical floor. The global scale fit accounts for the scale ambiguity of monocular reconstruction, while the displacement differences remove the dependence on the world-coordinate origin. 
Thus, the prior discourages changes in the relative coarse motion without preventing a common scale adjustment or refinements to the coarse and interpolated geometry. 
We weight this dimensionless loss by $w_{\mathrm{trans}}=0.05\,\mathcal{L}_{\mathrm{align}}^{\mathrm{start}}$, where $\mathcal{L}_{\mathrm{align}}^{\mathrm{start}}$ is the geometric alignment loss at the start of the second step; the weight remains fixed throughout that step.

\section{More Qualitative Results}
\label{sec:supp:qualitative}
\begingroup
\makeatletter
\setlength{\@fptop}{0pt}
\makeatother

\subsection{Depth estimation}
\label{sec:supp:qualitative:depth}

\Cref{fig:davis} shows depth interpolation on DAVIS~\citep{Perazzi16davis}, using the images at $t=0$ and $t=1$ and events simulated with vid2e.
The predictions at $t=\{0.25,0.5,0.75\}$ preserve the bear's silhouette and its separation from the background, extending the temporally coherent depth behavior of \cref{fig:bonn,fig:eds} to this additional sequence.

\Cref{fig:ablation:bonn} complements the Bonn comparison in \cref{fig:bonn}, with red boxes highlighting the moving person's silhouette and nearby occlusion boundaries.
Our method preserves more coherent boundaries across time, while the baselines introduce missing regions or distorted edges around the moving person.

\def\figWidth{0.18\linewidth}
\begin{figure}[t]
	\centering
    {\footnotesize
    \setlength{\tabcolsep}{0.5pt}
	\begin{tabular}{
	>{\centering\arraybackslash}m{0.3cm}
	>{\centering\arraybackslash}m{\figWidth}
	>{\centering\arraybackslash}m{\figWidth}
	>{\centering\arraybackslash}m{\figWidth}
	>{\centering\arraybackslash}m{\figWidth}
	>{\centering\arraybackslash}m{\figWidth}
	}

    \rotatebox{90}{\makecell{\scriptsize Frame}}
        &\imgbox{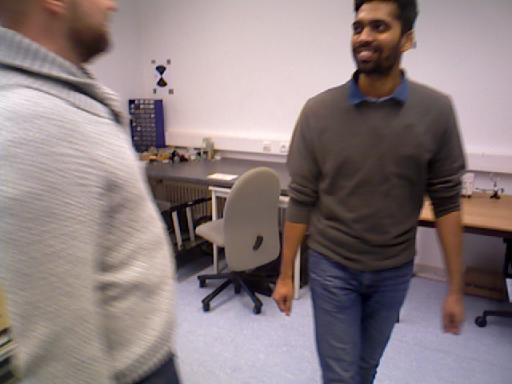}{(0.15,0.8) rectangle (0.3,0.99) (0.2,0.1) rectangle (0.4,0.7)}
        &\imgbox{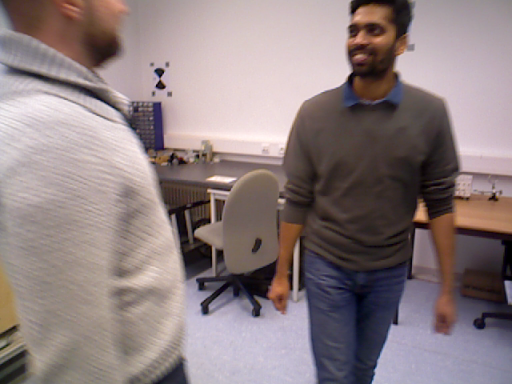}{(0.17,0.78) rectangle (0.32,0.99) (0.22,0.1) rectangle (0.42,0.7)}
        &\imgbox{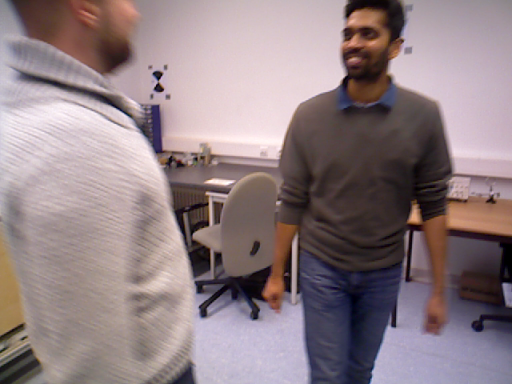}{(0.19,0.76) rectangle (0.34,0.99) (0.24,0.1) rectangle (0.44,0.7)}
        &\imgbox{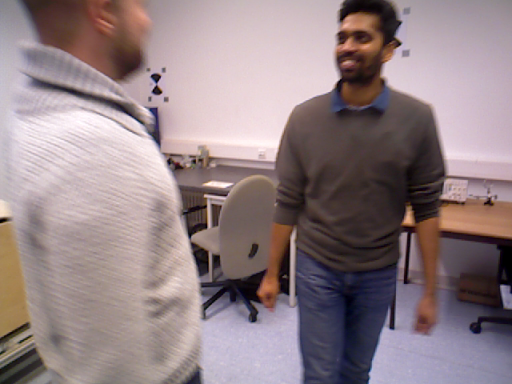}{(0.21,0.74) rectangle (0.36,0.99) (0.01,0.45) rectangle (0.1,0.9)}
        &\imgbox{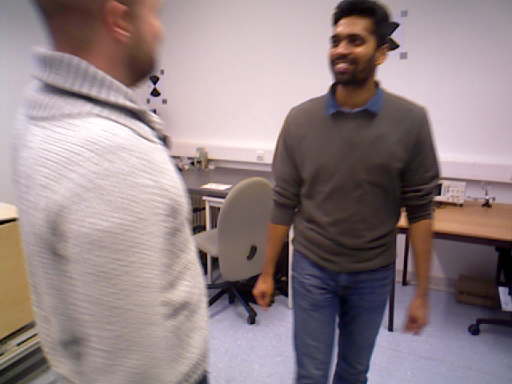}{(0.23,0.72) rectangle (0.38,0.99) (0.01,0.45) rectangle (0.12,0.9)}
	\\[-0.5ex]

    \rotatebox{90}{\makecell{\scriptsize RIFE}}
        &\imgbox{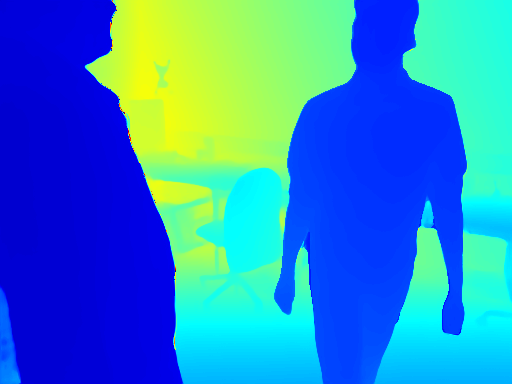}{(0.15,0.8) rectangle (0.3,0.99) (0.2,0.1) rectangle (0.4,0.7)}
        &\imgbox{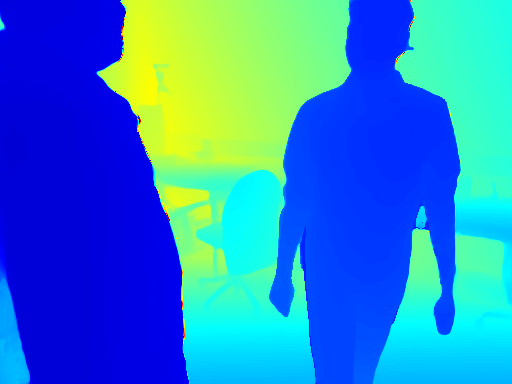}{(0.17,0.78) rectangle (0.32,0.99) (0.22,0.1) rectangle (0.42,0.7)}
        &\imgbox{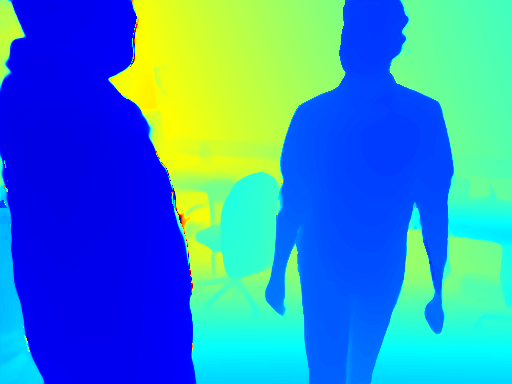}{(0.19,0.76) rectangle (0.34,0.99) (0.24,0.1) rectangle (0.44,0.7)}
        &\imgbox{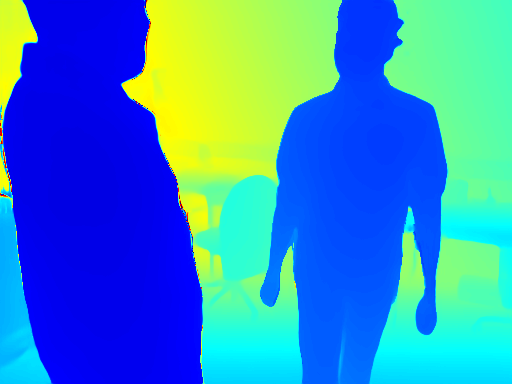}{(0.21,0.74) rectangle (0.36,0.99) (0.01,0.45) rectangle (0.1,0.9)}
        &\imgbox{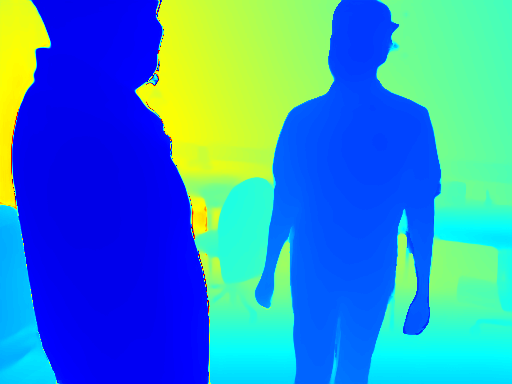}{(0.23,0.72) rectangle (0.38,0.99)  (0.01,0.45) rectangle (0.12,0.9)}
	\\[-0.5ex]

    \rotatebox{90}{\makecell{\scriptsize TimeLens}}
	&\imgbox{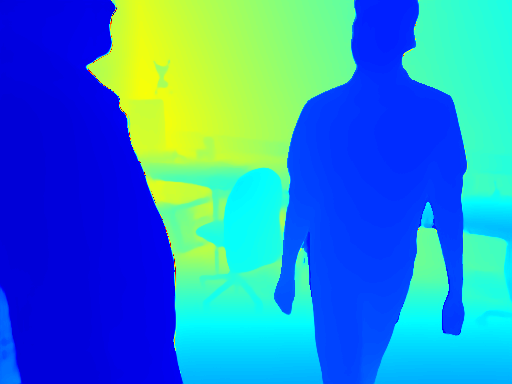}{(0.15,0.8) rectangle (0.3,0.99) (0.2,0.1) rectangle (0.4,0.7)}
        &\imgbox{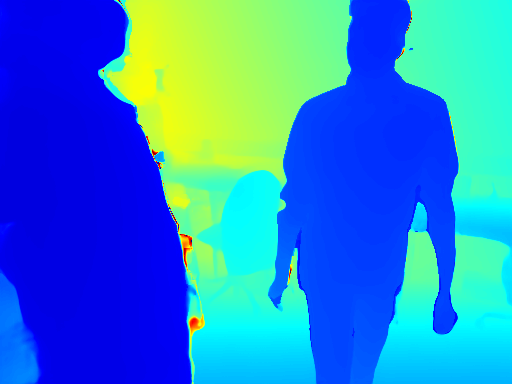}{(0.17,0.78) rectangle (0.32,0.99) (0.22,0.1) rectangle (0.42,0.7)}
        &\imgbox{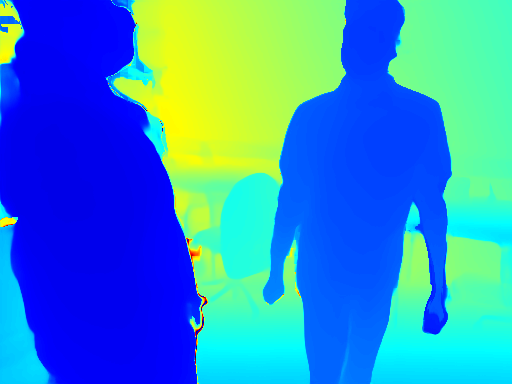}{(0.19,0.76) rectangle (0.34,0.99) (0.24,0.1) rectangle (0.44,0.7)}
        &\imgbox{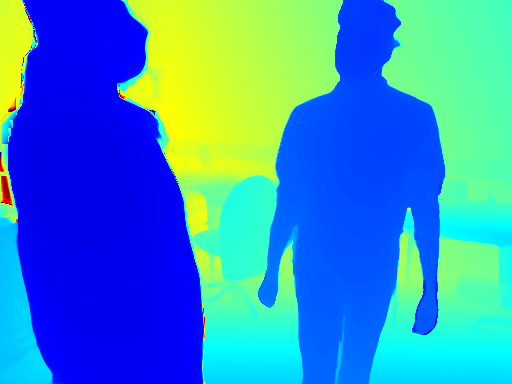}{(0.21,0.74) rectangle (0.36,0.99) (0.01,0.45) rectangle (0.1,0.9)}
        &\imgbox{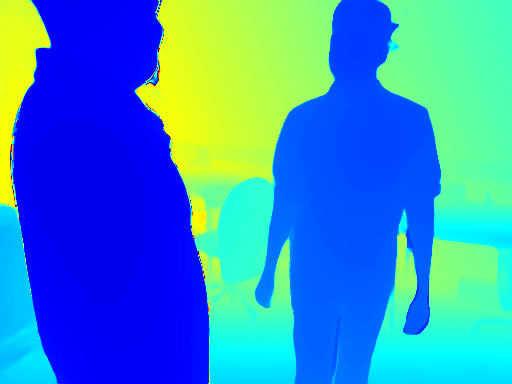}{(0.23,0.72) rectangle (0.38,0.99)  (0.01,0.45) rectangle (0.12,0.9)}
	\\[-0.5ex]

    \rotatebox{90}{\makecell{\scriptsize VDM-EVFI}}
	&\imgbox{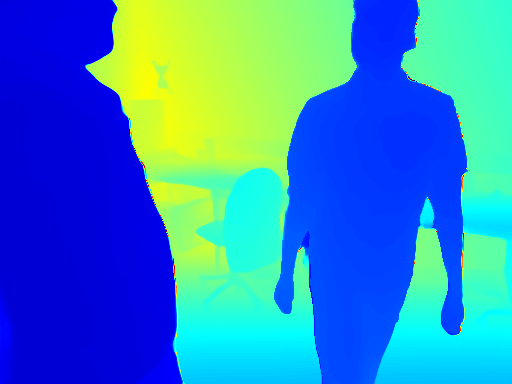}{(0.15,0.8) rectangle (0.3,0.99) (0.2,0.1) rectangle (0.4,0.7)}
        &\imgbox{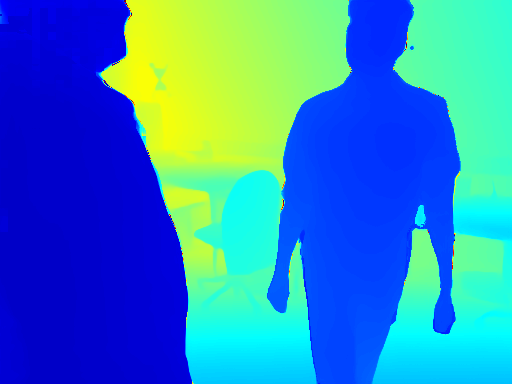}{(0.17,0.78) rectangle (0.32,0.99) (0.22,0.1) rectangle (0.42,0.7)}
        &\imgbox{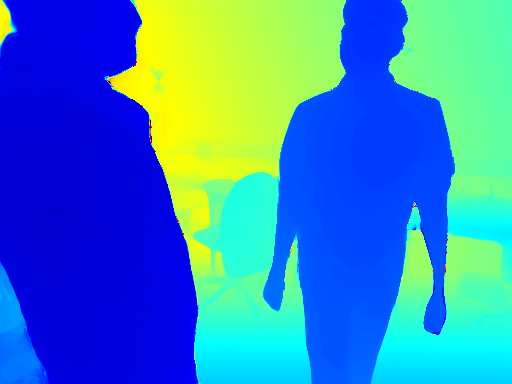}{(0.19,0.76) rectangle (0.34,0.99) (0.24,0.1) rectangle (0.44,0.7)}
        &\imgbox{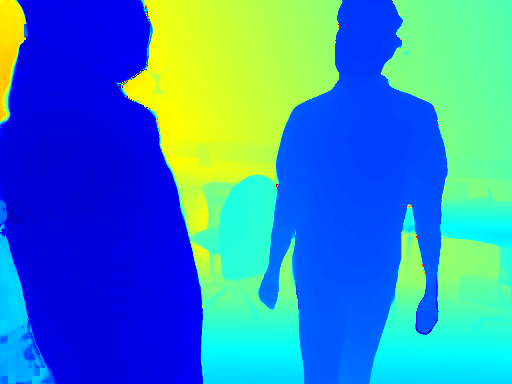}{(0.21,0.74) rectangle (0.36,0.99) (0.01,0.45) rectangle (0.1,0.9)}
        &\imgbox{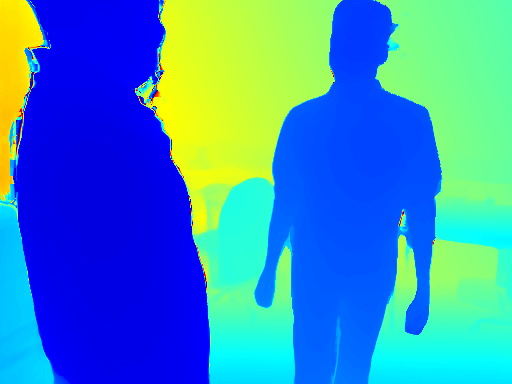}{(0.23,0.72) rectangle (0.38,0.99)  (0.01,0.45) rectangle (0.12,0.9)}
	\\[-0.5ex]

    \rotatebox{90}{\makecell{\scriptsize Ours}}
	&\imgbox{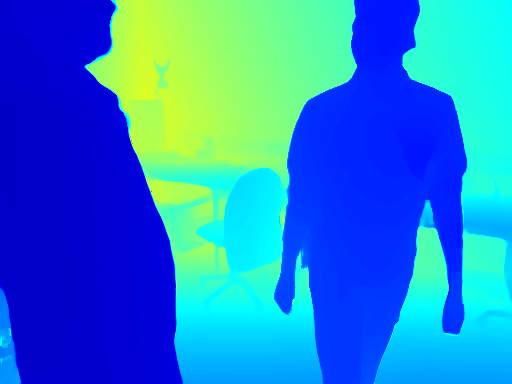}{(0.15,0.8) rectangle (0.3,0.99) (0.2,0.1) rectangle (0.4,0.7)}
        &\imgbox{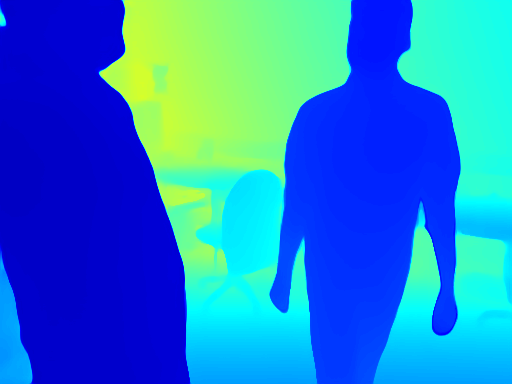}{(0.17,0.78) rectangle (0.32,0.99) (0.22,0.1) rectangle (0.42,0.7)}
        &\imgbox{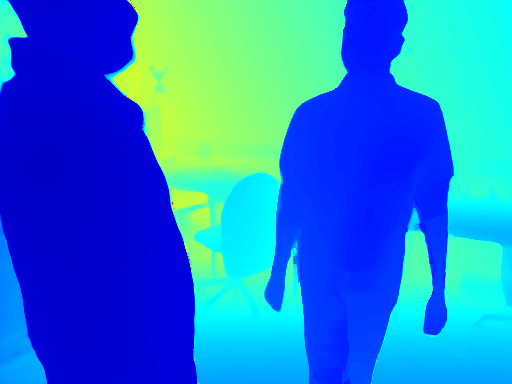}{(0.19,0.76) rectangle (0.34,0.99) (0.24,0.1) rectangle (0.44,0.7)}
        &\imgbox{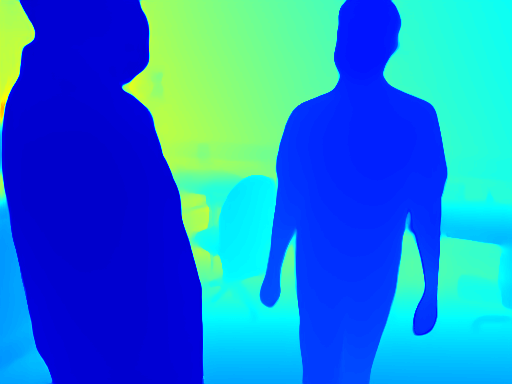}{(0.21,0.74) rectangle (0.36,0.99) (0.01,0.45) rectangle (0.1,0.9)}
        &\imgbox{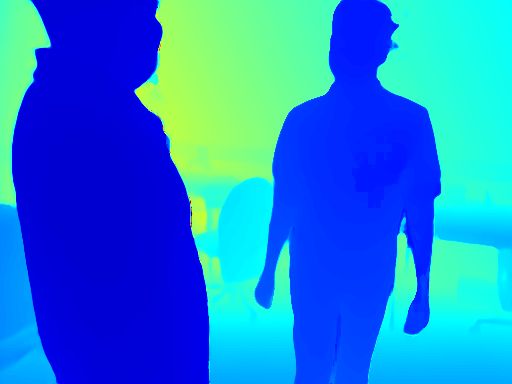}{(0.23,0.72) rectangle (0.38,0.99) (0.01,0.45) rectangle (0.12,0.9)}
	\\[-0.5ex]

        & \scriptsize $t = 0$
        & \scriptsize $t = 0.25$
        & \scriptsize $t = 0.5$
        & \scriptsize $t = 0.75$
        & \scriptsize $t = 1$
        \\[-0.5ex]

	\end{tabular}
	}
    \caption{Qualitative depth comparison on the Bonn \texttt{crowd2} sequence (skip = 3). \label{fig:ablation:bonn}}
\end{figure}

\newcommand{\edsdepthpanel}[3]{%
  \imgbox{images/qualitative_depthmaps/depthmaps_#1/#2_#1/t#3.png}{\edsdepthboxes}}

\newcommand{\edsdepthcomparison}[4]{%
  \begin{figure}[htbp]
    \centering
    \def\edsdepthwidth{0.18\linewidth}
    \def\edsdepthboxes{#4}
    {\footnotesize\setlength{\tabcolsep}{0.5pt}
    \begin{tabular}{
      >{\centering\arraybackslash}m{0.3cm}
      *{5}{>{\centering\arraybackslash}m{\edsdepthwidth}}
    }
      \rotatebox{90}{\scriptsize Frame}
        & \edsdepthpanel{#1}{frames}{0}
        & \edsdepthpanel{#1}{frames}{0-25}
        & \edsdepthpanel{#1}{frames}{0-50}
        & \edsdepthpanel{#1}{frames}{0-75}
        & \edsdepthpanel{#1}{frames}{1} \\[-0.5ex]
      \rotatebox{90}{\scriptsize RIFE}
        & \edsdepthpanel{#1}{rife}{0}
        & \edsdepthpanel{#1}{rife}{0-25}
        & \edsdepthpanel{#1}{rife}{0-50}
        & \edsdepthpanel{#1}{rife}{0-75}
        & \edsdepthpanel{#1}{rife}{1} \\[-0.5ex]
      \rotatebox{90}{\scriptsize TimeLens}
        & \edsdepthpanel{#1}{timelens}{0}
        & \edsdepthpanel{#1}{timelens}{0-25}
        & \edsdepthpanel{#1}{timelens}{0-50}
        & \edsdepthpanel{#1}{timelens}{0-75}
        & \edsdepthpanel{#1}{timelens}{1} \\[-0.5ex]
      \rotatebox{90}{\scriptsize VDM-EVFI}
        & \edsdepthpanel{#1}{vdm}{0}
        & \edsdepthpanel{#1}{vdm}{0-25}
        & \edsdepthpanel{#1}{vdm}{0-50}
        & \edsdepthpanel{#1}{vdm}{0-75}
        & \edsdepthpanel{#1}{vdm}{1} \\[-0.5ex]
      \rotatebox{90}{\scriptsize Ours}
        & \edsdepthpanel{#1}{interp3r}{0}
        & \edsdepthpanel{#1}{interp3r}{0-25}
        & \edsdepthpanel{#1}{interp3r}{0-50}
        & \edsdepthpanel{#1}{interp3r}{0-75}
        & \edsdepthpanel{#1}{interp3r}{1} \\[-0.5ex]
        & \scriptsize $t=0$
        & \scriptsize $t=0.25$
        & \scriptsize $t=0.5$
        & \scriptsize $t=0.75$
        & \scriptsize $t=1$ \\
    \end{tabular}}
    \caption{Qualitative depth comparison on the EDS \texttt{#2} sequence.
    }
    \label{#3}
  \end{figure}}

\Cref{fig:depth:eds-peanuts} extends the real-event depth example in \cref{fig:eds} with baseline comparisons.
Within the red boxes, our method preserves the small objects and their boundaries more consistently across interpolated timestamps, while the baselines show changes in shape or local depth.
\edsdepthcomparison{01_peanuts_light_200_299_s1}{01\_peanuts\_light}{fig:depth:eds-peanuts}{%
  (0.31,0.42) rectangle (0.51,0.58)
  (0.61,0.18) rectangle (0.86,0.41)}

In \cref{fig:depth:eds-rocket}, the red box highlights fine scene structure behind the foreground objects.
Our depth maps retain a clearer separation of these structures from the background across time, supporting the generalization to real images and events already observed in previous figures.
\edsdepthcomparison{02_rocket_earth_light_500_599_s1}{02\_rocket\_earth\_light}{fig:depth:eds-rocket}{%
  (0.32,0.70) rectangle (0.67,0.90)}

\subsection{Camera pose estimation}
\label{sec:supp:qualitative:pose}
In \cref{fig:traj:ecd-boxes}, our method better preserves the curved trajectory of the ECD \texttt{boxes\_6dof} sequence.
At larger skips, the baselines exhibit sharp jumps and intersecting segments, while our estimates retain a more coherent path around the scene.
\begin{figure}[t]
  \centering
  \def\trajectoryheight{0.20\textheight}%
  {\footnotesize\setlength{\tabcolsep}{0.5pt}%
  \begin{tabular}{@{}>{\centering\arraybackslash}m{0.3cm}*{2}{>{\centering\arraybackslash}m{0.48\linewidth}}@{}}
    \rotatebox{90}{\scriptsize RIFE}
      & \includegraphics[trim=0bp 0bp 0bp 27bp,clip,width=\linewidth,height=\trajectoryheight,keepaspectratio]{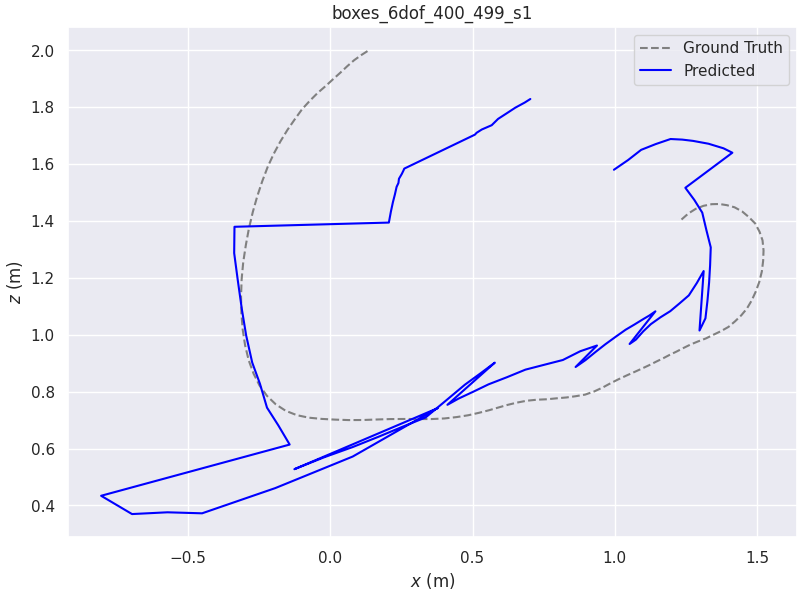} & \includegraphics[trim=0bp 0bp 0bp 29bp,clip,width=\linewidth,height=\trajectoryheight,keepaspectratio]{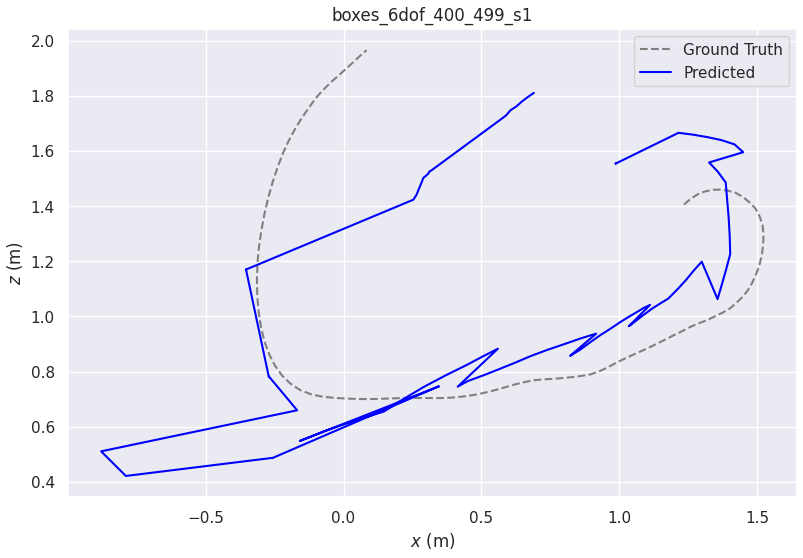} \\[-0.5ex]
    \rotatebox{90}{\scriptsize TimeLens}
      & \includegraphics[trim=0bp 0bp 0bp 27bp,clip,width=\linewidth,height=\trajectoryheight,keepaspectratio]{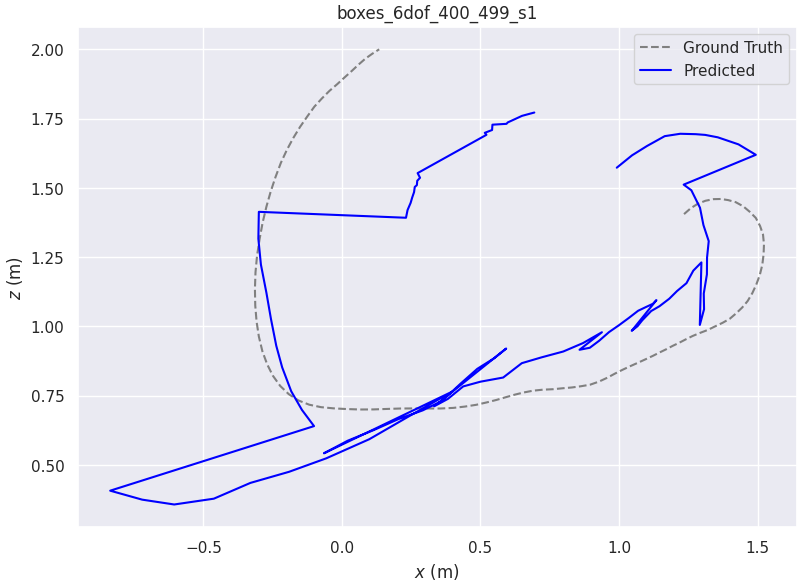} & \includegraphics[trim=0bp 0bp 0bp 29bp,clip,width=\linewidth,height=\trajectoryheight,keepaspectratio]{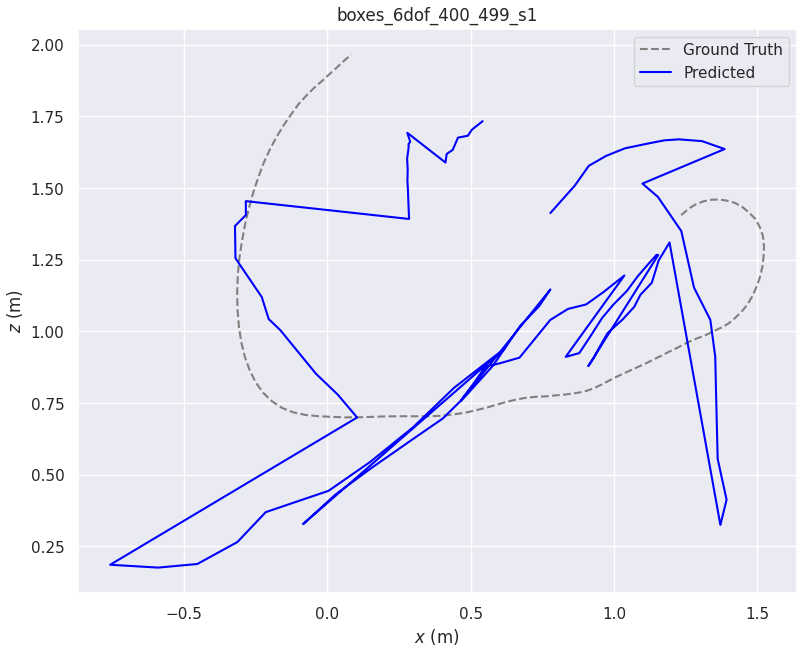} \\[-0.5ex]
    \rotatebox{90}{\scriptsize VDM-EVFI}
      & \includegraphics[trim=0bp 0bp 0bp 29bp,clip,width=\linewidth,height=\trajectoryheight,keepaspectratio]{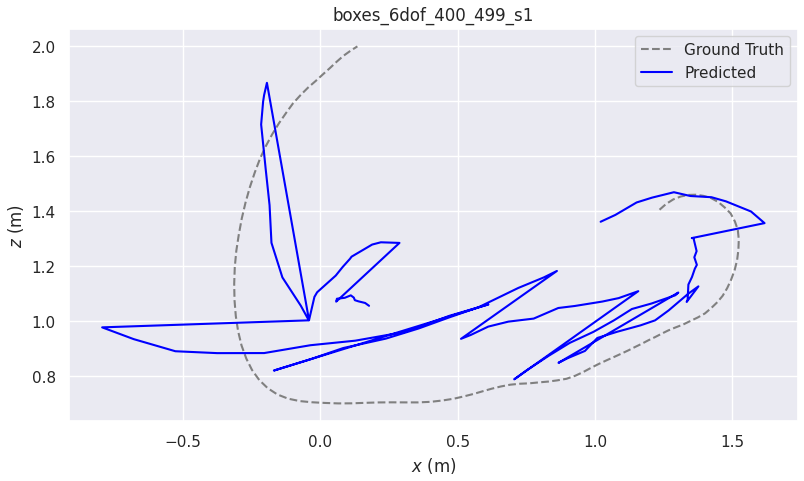} & \includegraphics[trim=0bp 0bp 0bp 29bp,clip,width=\linewidth,height=\trajectoryheight,keepaspectratio]{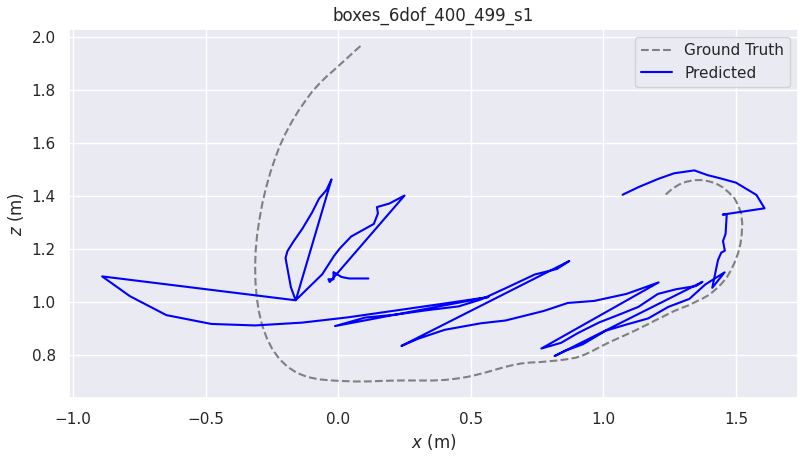} \\[-0.5ex]
    \rotatebox{90}{\scriptsize Ours}
      & \includegraphics[trim=0bp 0bp 0bp 28bp,clip,width=\linewidth,height=\trajectoryheight,keepaspectratio]{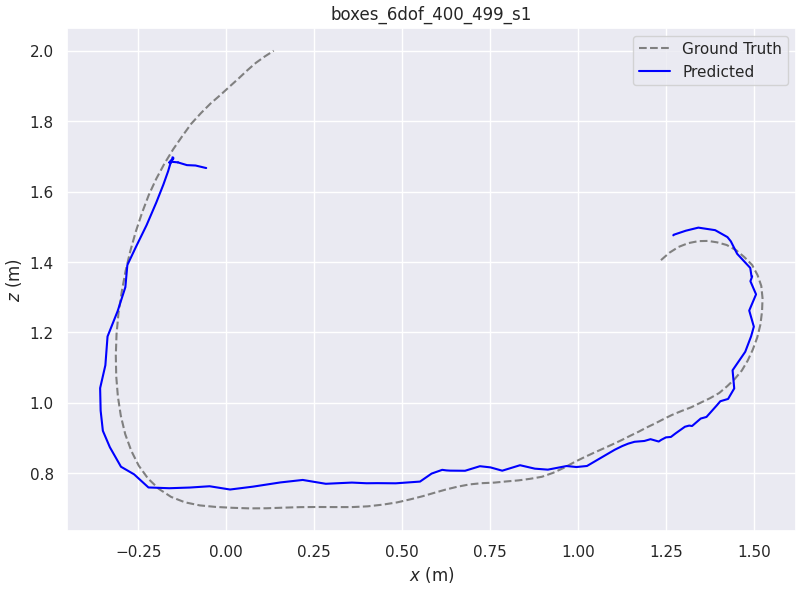} & \includegraphics[trim=0bp 0bp 0bp 29bp,clip,width=\linewidth,height=\trajectoryheight,keepaspectratio]{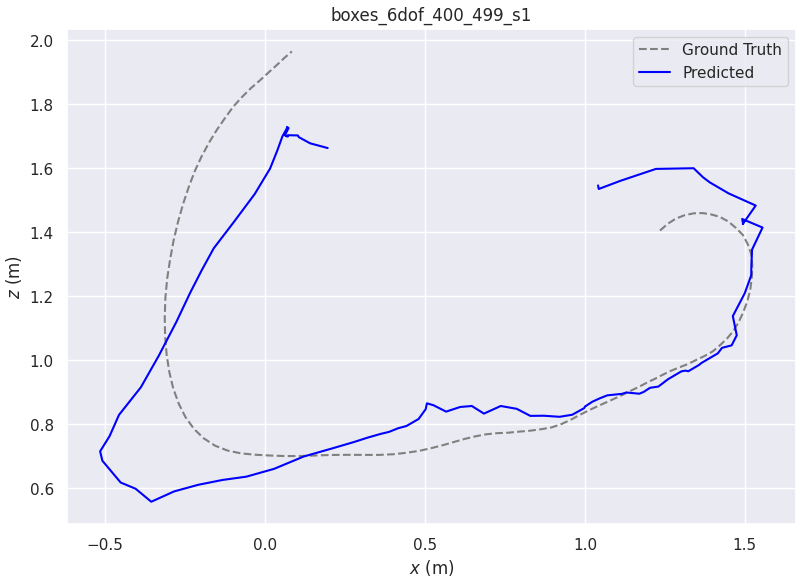} \\[-0.5ex]
      & \scriptsize $\mathrm{skip}=1$ & \scriptsize $\mathrm{skip}=3$ \\
  \end{tabular}}
  \caption{Camera trajectories on the ECD \texttt{boxes\_6dof} sequence.
  Dashed gray lines indicate ground truth; solid blue lines indicate estimates.}
  \label{fig:traj:ecd-boxes}
\end{figure}
\begin{figure}[t]
  \ContinuedFloat
  \centering
  \def\trajectoryheight{0.20\textheight}%
  {\footnotesize\setlength{\tabcolsep}{0.5pt}%
  \begin{tabular}{@{}>{\centering\arraybackslash}m{0.3cm}*{2}{>{\centering\arraybackslash}m{0.48\linewidth}}@{}}
    \rotatebox{90}{\scriptsize RIFE}
      & \includegraphics[trim=0bp 0bp 0bp 29bp,clip,width=\linewidth,height=\trajectoryheight,keepaspectratio]{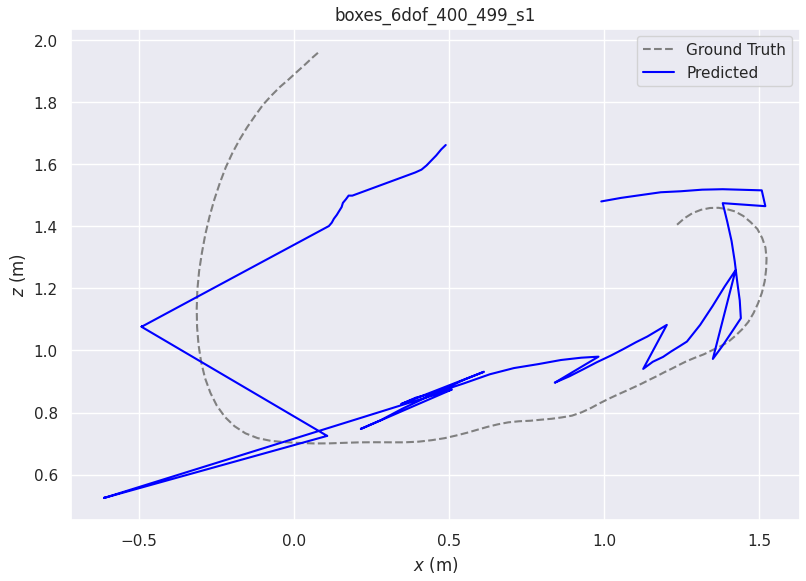} & \includegraphics[trim=0bp 0bp 0bp 30bp,clip,width=\linewidth,height=\trajectoryheight,keepaspectratio]{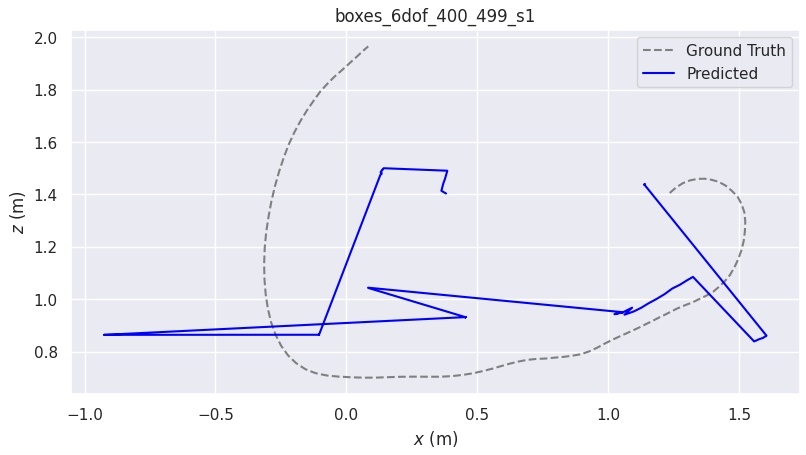} \\[-0.5ex]
    \rotatebox{90}{\scriptsize TimeLens}
      & \includegraphics[trim=0bp 0bp 0bp 28bp,clip,width=\linewidth,height=\trajectoryheight,keepaspectratio]{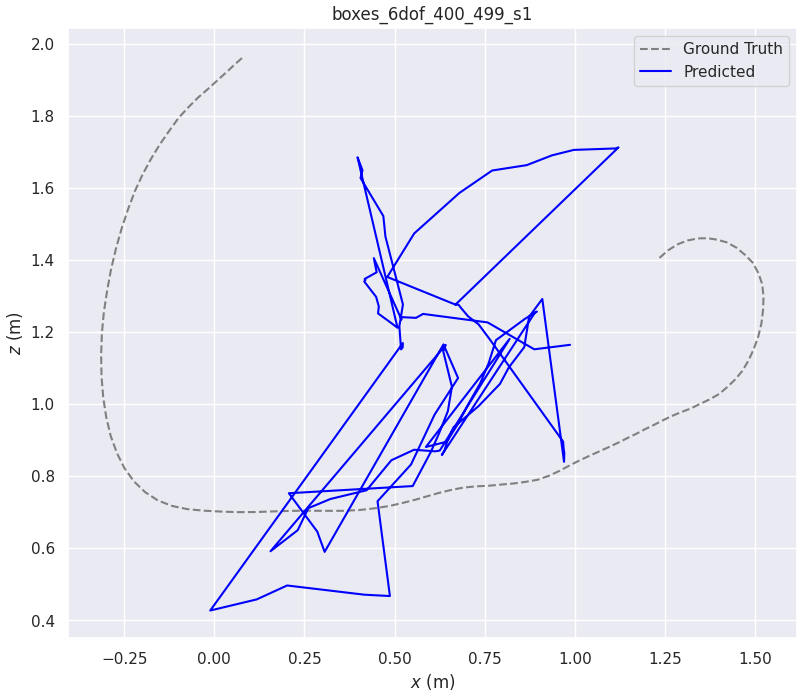} & \includegraphics[trim=0bp 0bp 0bp 27bp,clip,width=\linewidth,height=\trajectoryheight,keepaspectratio]{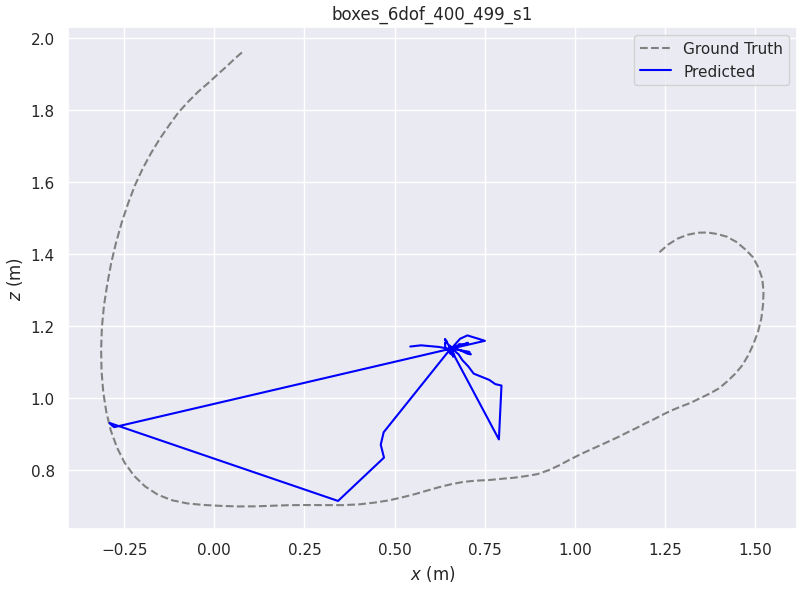} \\[-0.5ex]
    \rotatebox{90}{\scriptsize VDM-EVFI}
      & \includegraphics[trim=0bp 0bp 0bp 29bp,clip,width=\linewidth,height=\trajectoryheight,keepaspectratio]{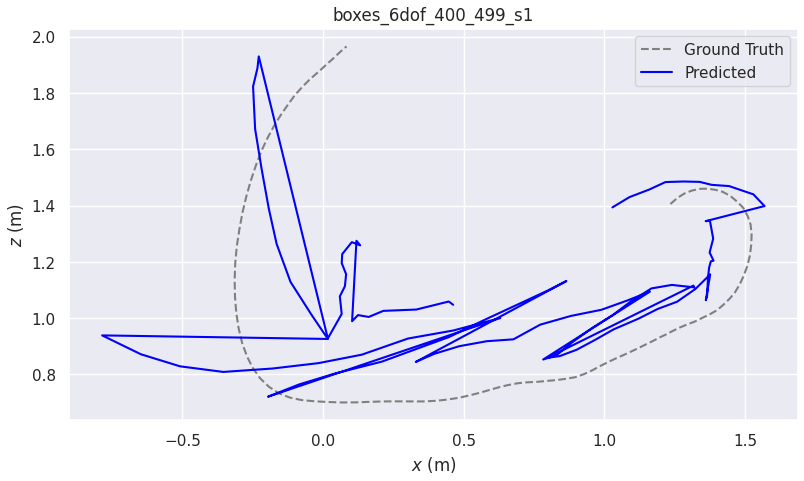} & \includegraphics[trim=0bp 0bp 0bp 28bp,clip,width=\linewidth,height=\trajectoryheight,keepaspectratio]{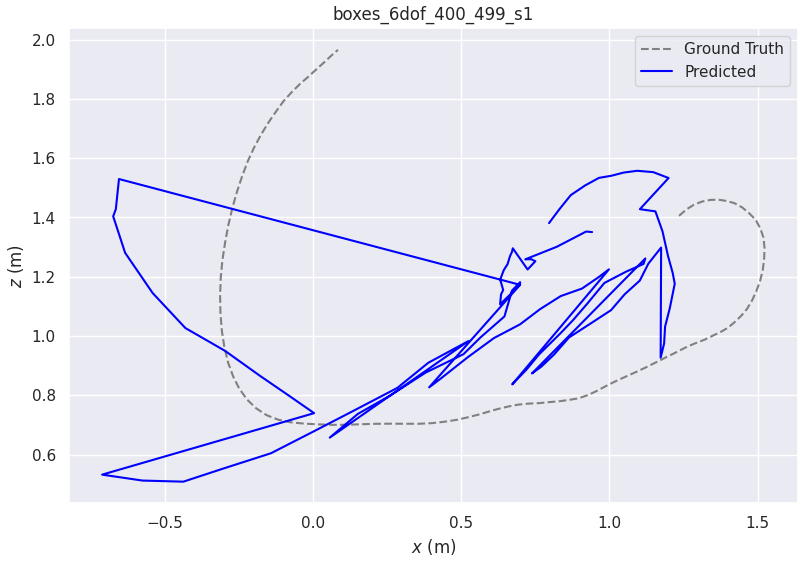} \\[-0.5ex]
    \rotatebox{90}{\scriptsize Ours}
      & \includegraphics[trim=0bp 0bp 0bp 29bp,clip,width=\linewidth,height=\trajectoryheight,keepaspectratio]{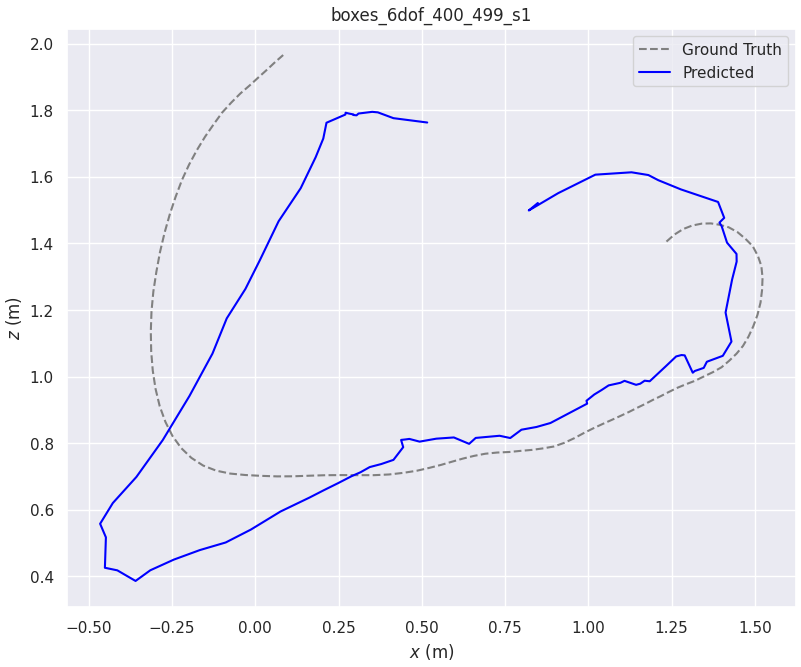} & \includegraphics[trim=0bp 0bp 0bp 29bp,clip,width=\linewidth,height=\trajectoryheight,keepaspectratio]{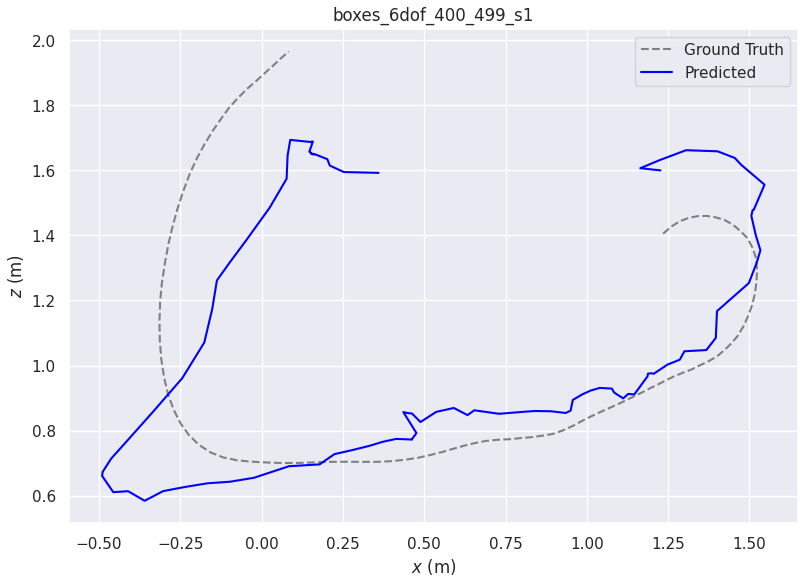} \\[-0.5ex]
      & \scriptsize $\mathrm{skip}=7$ & \scriptsize $\mathrm{skip}=15$ \\
  \end{tabular}}
  \caption[]{Camera trajectories on ECD \texttt{boxes\_6dof} (continued).}
\end{figure}

\Cref{fig:traj:ecd-dynamic} illustrates pose estimation in a scene with independent object motion.
At skip $=7$, our trajectory follows the groundtruth turn smoothly, with fewer abrupt directional changes than the baseline estimates.
\begin{figure}[t]
  \centering
  \def\trajectoryheight{0.098\textheight}%
  {\footnotesize\setlength{\tabcolsep}{0.5pt}%
  \begin{tabular}{@{}>{\centering\arraybackslash}m{0.3cm}*{2}{>{\centering\arraybackslash}m{0.48\linewidth}}@{}}
    \rotatebox{90}{\scriptsize RIFE}
      & \includegraphics[trim=0bp 0bp 0bp 29bp,clip,width=\linewidth,height=\trajectoryheight,keepaspectratio]{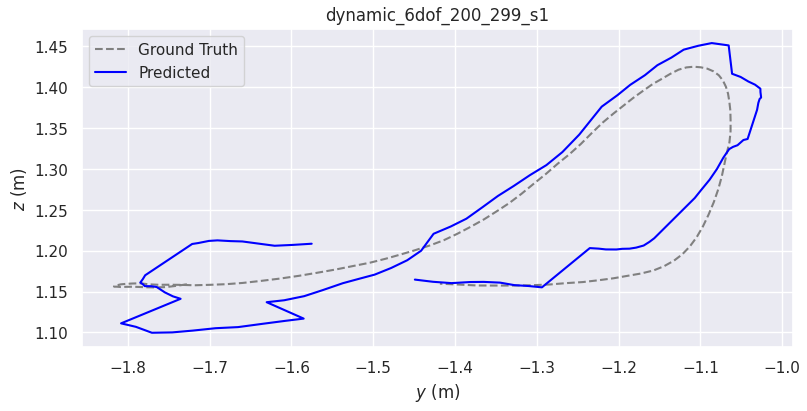} & \includegraphics[trim=0bp 0bp 0bp 30bp,clip,width=\linewidth,height=\trajectoryheight,keepaspectratio]{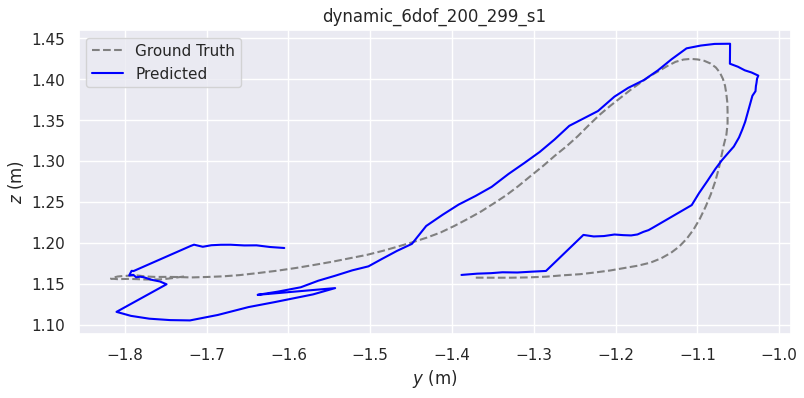} \\[-0.5ex]
    \rotatebox{90}{\scriptsize TimeLens}
      & \includegraphics[trim=0bp 0bp 0bp 27bp,clip,width=\linewidth,height=\trajectoryheight,keepaspectratio]{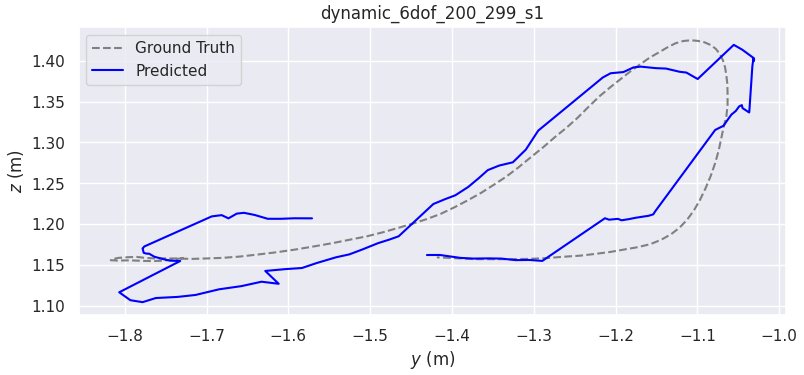} & \includegraphics[trim=0bp 0bp 0bp 29bp,clip,width=\linewidth,height=\trajectoryheight,keepaspectratio]{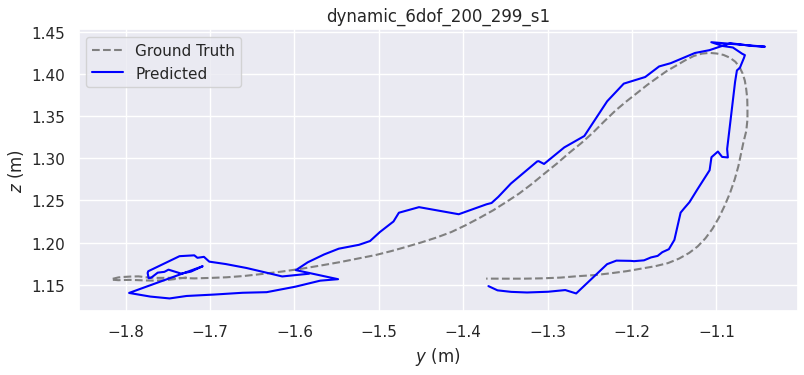} \\[-0.5ex]
    \rotatebox{90}{\scriptsize VDM-EVFI}
      & \includegraphics[trim=0bp 0bp 0bp 27bp,clip,width=\linewidth,height=\trajectoryheight,keepaspectratio]{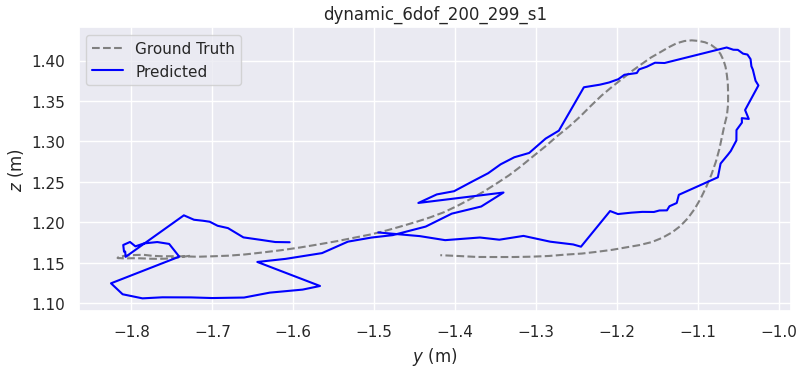} & \includegraphics[trim=0bp 0bp 0bp 28bp,clip,width=\linewidth,height=\trajectoryheight,keepaspectratio]{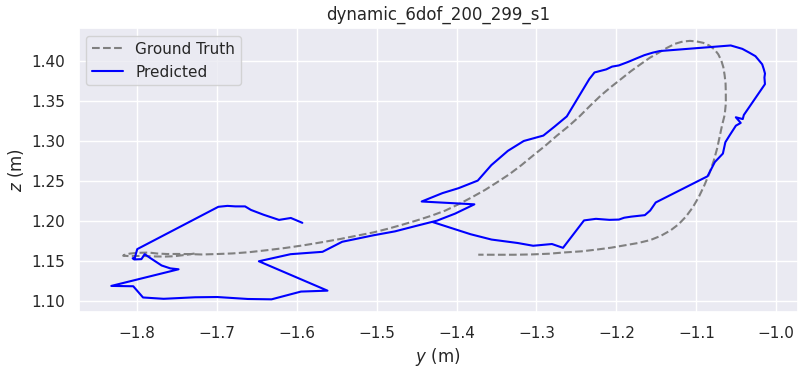} \\[-0.5ex]
    \rotatebox{90}{\scriptsize Ours}
      & \includegraphics[trim=0bp 0bp 0bp 29bp,clip,width=\linewidth,height=\trajectoryheight,keepaspectratio]{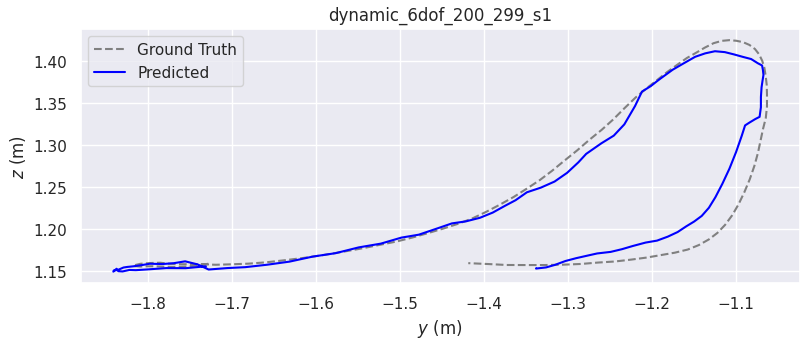} & \includegraphics[trim=0bp 0bp 0bp 29bp,clip,width=\linewidth,height=\trajectoryheight,keepaspectratio]{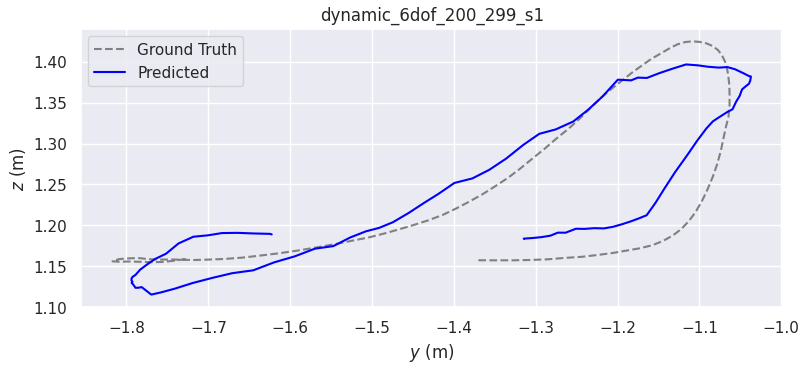} \\[-0.5ex]
      & \scriptsize $\mathrm{skip}=1$ & \scriptsize $\mathrm{skip}=3$ \\
  \end{tabular}}
  \par\medskip
  \def\trajectoryheight{0.098\textheight}%
  {\footnotesize\setlength{\tabcolsep}{0.5pt}%
  \begin{tabular}{@{}>{\centering\arraybackslash}m{0.3cm}*{2}{>{\centering\arraybackslash}m{0.48\linewidth}}@{}}
    \rotatebox{90}{\scriptsize RIFE}
      & \includegraphics[trim=0bp 0bp 0bp 30bp,clip,width=\linewidth,height=\trajectoryheight,keepaspectratio]{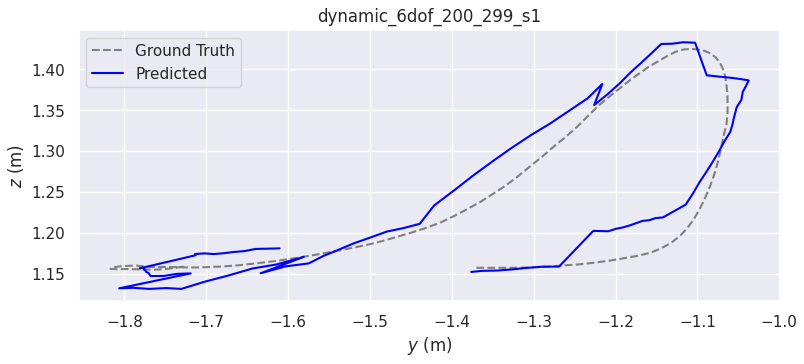} & \includegraphics[trim=0bp 0bp 0bp 27bp,clip,width=\linewidth,height=\trajectoryheight,keepaspectratio]{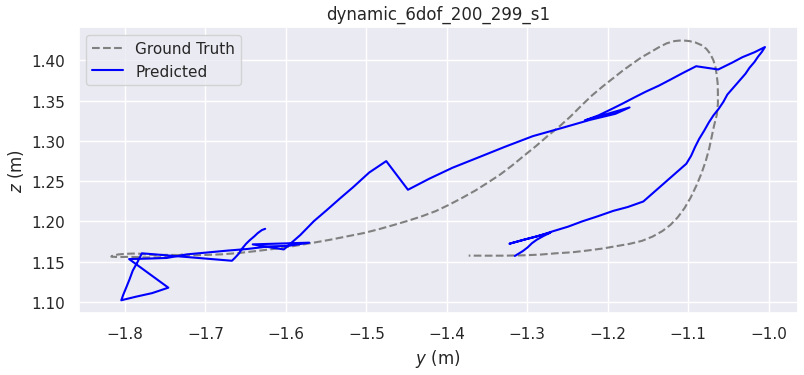} \\[-0.5ex]
    \rotatebox{90}{\scriptsize TimeLens}
      & \includegraphics[trim=0bp 0bp 0bp 27bp,clip,width=\linewidth,height=\trajectoryheight,keepaspectratio]{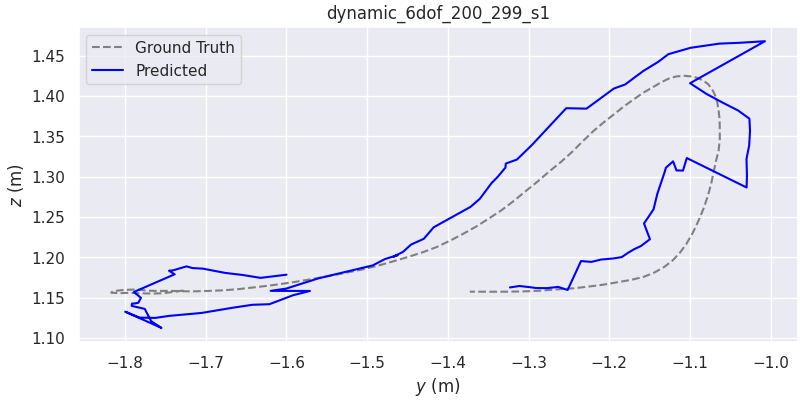} & \includegraphics[trim=0bp 0bp 0bp 28bp,clip,width=\linewidth,height=\trajectoryheight,keepaspectratio]{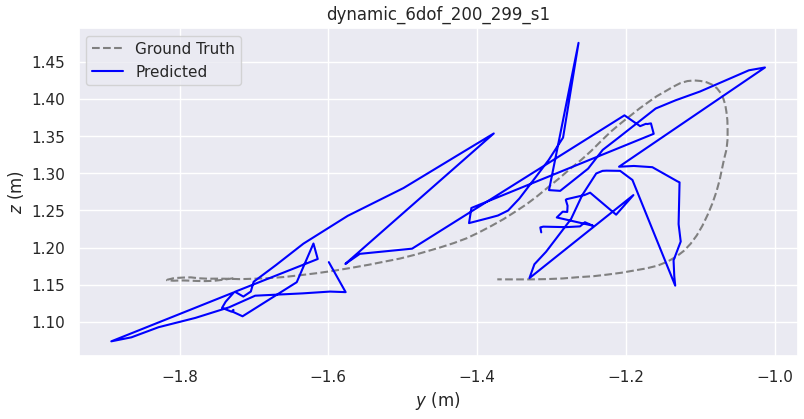} \\[-0.5ex]
    \rotatebox{90}{\scriptsize VDM-EVFI}
      & \includegraphics[trim=0bp 0bp 0bp 28bp,clip,width=\linewidth,height=\trajectoryheight,keepaspectratio]{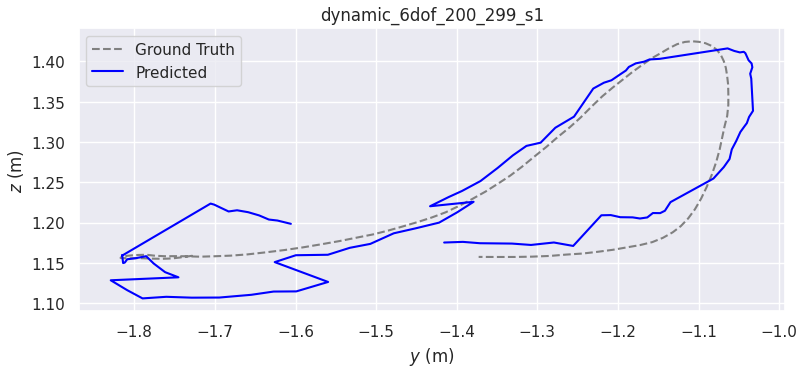} & \includegraphics[trim=0bp 0bp 0bp 29bp,clip,width=\linewidth,height=\trajectoryheight,keepaspectratio]{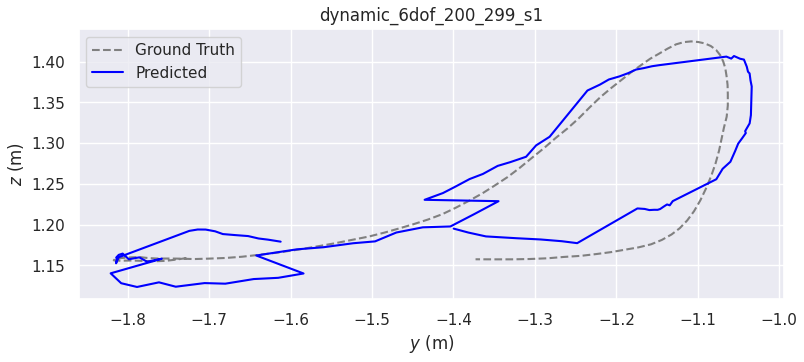} \\[-0.5ex]
    \rotatebox{90}{\scriptsize Ours}
      & \includegraphics[trim=0bp 0bp 0bp 31bp,clip,width=\linewidth,height=\trajectoryheight,keepaspectratio]{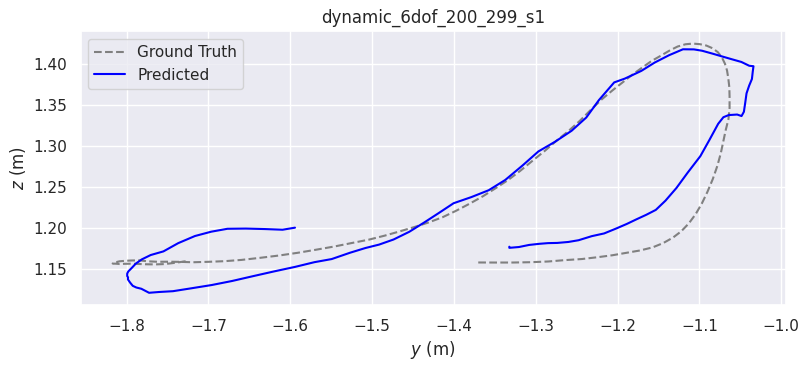} & \includegraphics[trim=0bp 0bp 0bp 28bp,clip,width=\linewidth,height=\trajectoryheight,keepaspectratio]{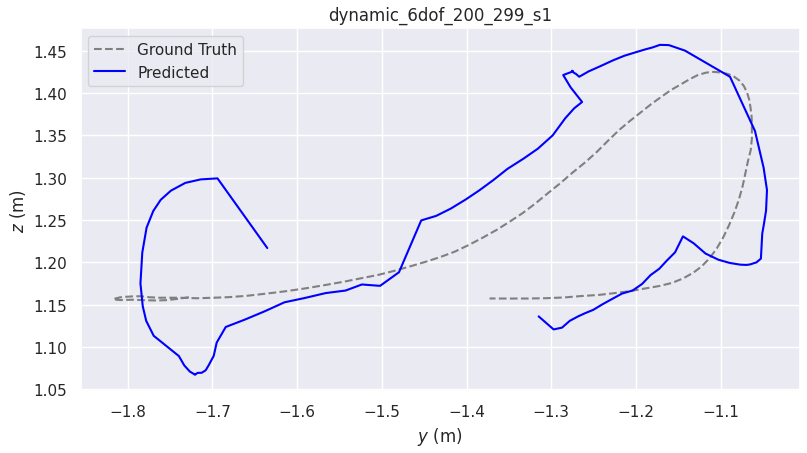} \\[-0.5ex]
      & \scriptsize $\mathrm{skip}=7$ & \scriptsize $\mathrm{skip}=15$ \\
  \end{tabular}}
  \caption{Camera trajectories on the ECD \texttt{dynamic\_6dof} sequence.
  Dashed gray lines indicate ground truth; solid blue lines indicate estimates.}
  \label{fig:traj:ecd-dynamic}
\end{figure}

\Cref{fig:traj:eds-peanuts} shows that our estimated trajectory follows the overall groundtruth path on EDS even with sparse input frames.
At skip $=15$, it remains smoother than the abrupt local changes visible in the interpolation-based baselines.
\begin{figure}[t]
  \centering
  \def\trajectoryheight{0.08\textheight}%
  {\footnotesize\setlength{\tabcolsep}{0.5pt}%
  \begin{tabular}{@{}>{\centering\arraybackslash}m{0.3cm}*{2}{>{\centering\arraybackslash}m{0.48\linewidth}}@{}}
    \rotatebox{90}{\scriptsize RIFE}
      & \includegraphics[trim=0bp 0bp 0bp 28bp,clip,width=\linewidth,height=\trajectoryheight,keepaspectratio]{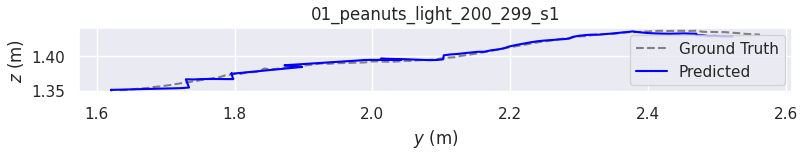} & \includegraphics[trim=0bp 0bp 0bp 28bp,clip,width=\linewidth,height=\trajectoryheight,keepaspectratio]{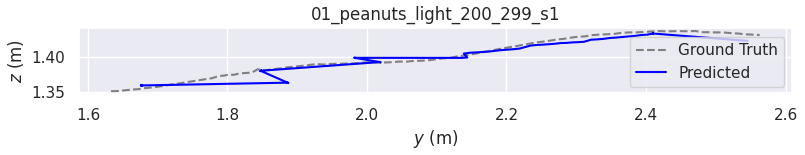} \\[-0.5ex]
    \rotatebox{90}{\scriptsize TimeLens}
      & \includegraphics[trim=0bp 0bp 0bp 27bp,clip,width=\linewidth,height=\trajectoryheight,keepaspectratio]{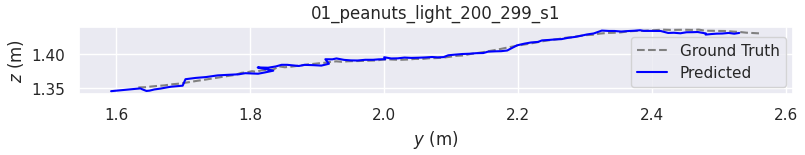} & \includegraphics[trim=0bp 0bp 0bp 25bp,clip,width=\linewidth,height=\trajectoryheight,keepaspectratio]{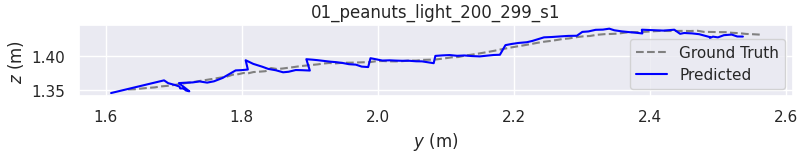} \\[-0.5ex]
    \rotatebox{90}{\scriptsize VDM-EVFI}
      & \includegraphics[trim=0bp 0bp 0bp 28bp,clip,width=\linewidth,height=\trajectoryheight,keepaspectratio]{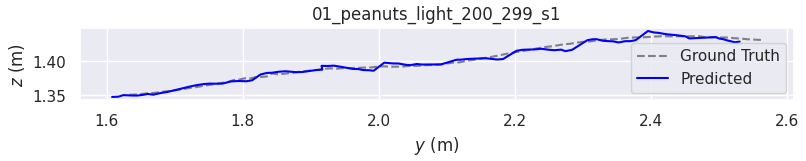} & \includegraphics[trim=0bp 0bp 0bp 29bp,clip,width=\linewidth,height=\trajectoryheight,keepaspectratio]{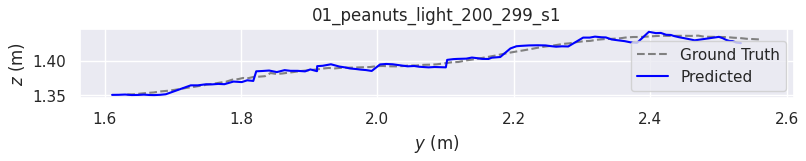} \\[-0.5ex]
    \rotatebox{90}{\scriptsize Ours}
      & \includegraphics[trim=0bp 0bp 0bp 29bp,clip,width=\linewidth,height=\trajectoryheight,keepaspectratio]{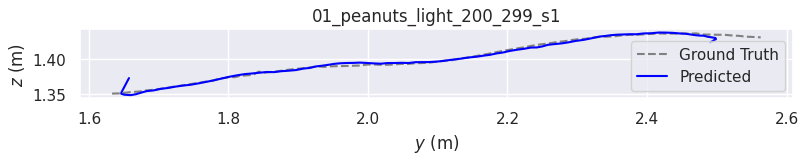} & \includegraphics[trim=0bp 0bp 0bp 27bp,clip,width=\linewidth,height=\trajectoryheight,keepaspectratio]{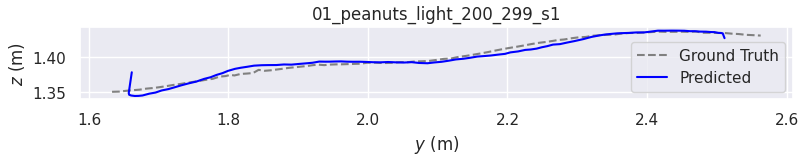} \\[-0.5ex]
      & \scriptsize $\mathrm{skip}=7$ & \scriptsize $\mathrm{skip}=15$ \\
  \end{tabular}}
  \caption{Camera trajectories on the EDS \texttt{01\_peanuts\_light} sequence.
  Dashed gray lines indicate ground truth; solid blue lines indicate estimates.}
  \label{fig:traj:eds-peanuts}
\end{figure}

\subsection{Visualization of Camera Poses and 3D Points}
\label{sec:supp:qualitative:points}

In \cref{fig:pointmaps:ecd-dynamic}, our method reconstructs a smoother camera path while retaining the spatial layout of the indoor scene.
Compared with the baselines, the camera frusta exhibit fewer abrupt changes in position and orientation, consistent with the trajectory comparison in \cref{fig:traj:ecd-dynamic}.
\def\pointmaprowheight{0.20001\linewidth}
\pointmapcomparison{3D reconstruction on the ECD \texttt{dynamic\_6dof} sequence (skip = 3).}%
  {{\includegraphics[trim=30bp 5bp 0bp 0bp,clip,height=\pointmaprowheight]{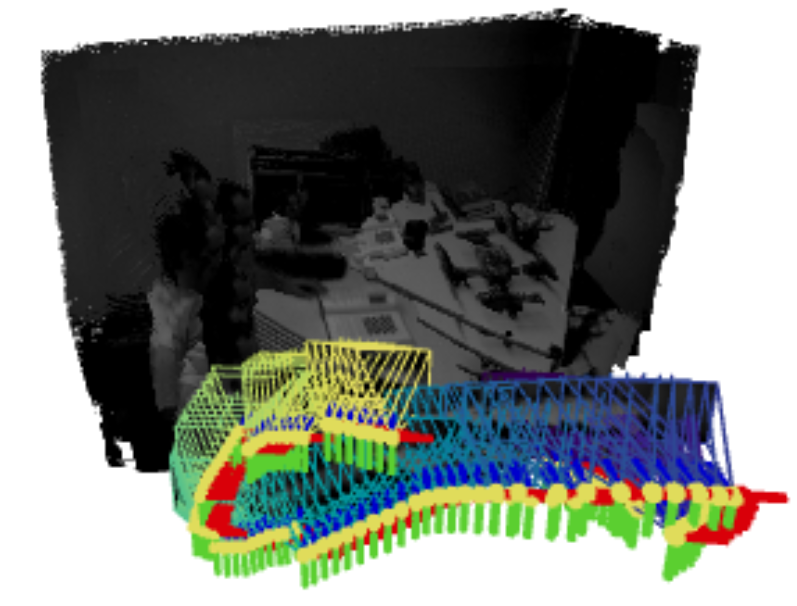}}}%
  {{\includegraphics[trim=57bp 3bp 11bp 12bp,clip,height=\pointmaprowheight]{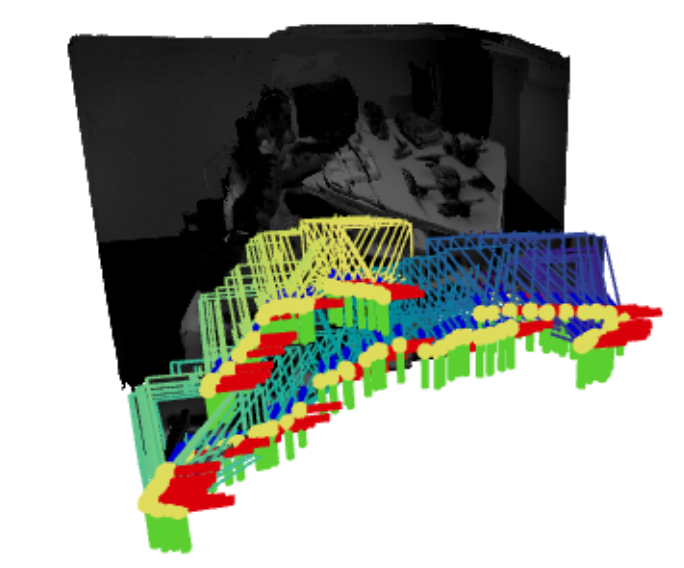}}}%
  {{\includegraphics[trim=67bp 0bp 24bp 13bp,clip,height=\pointmaprowheight]{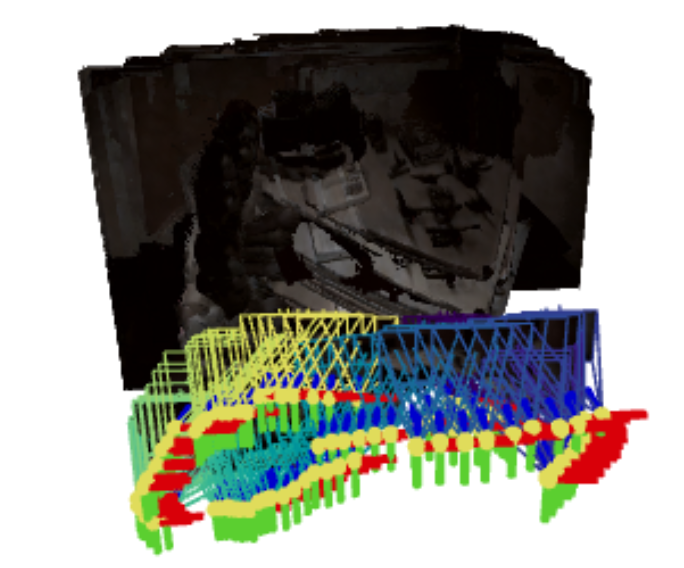}}}%
  {{\includegraphics[trim=85bp 47bp 32bp 52bp,clip,height=\pointmaprowheight]{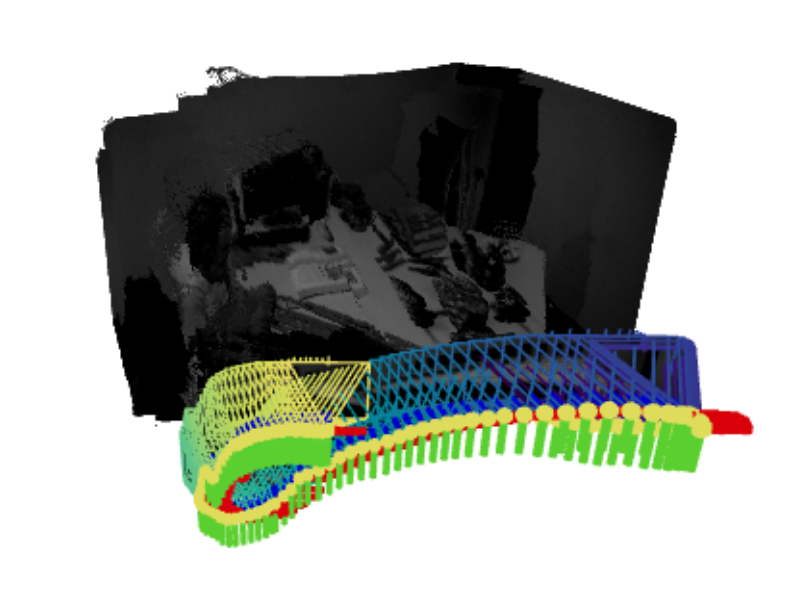}}}%
  {fig:pointmaps:ecd-dynamic}

In \cref{fig:pointmaps:dsec09}, our reconstruction better preserves the relative layout of the street and surrounding buildings.
The camera frusta form a more regular progression through the scene than the uneven or distorted paths visible in the baselines.
\def\pointmaprowheight{0.17209\linewidth}
\pointmapcomparison{3D reconstruction on the \texttt{dsec\_zurich\_city\_09d} sequence (skip = 3).}%
  {{\includegraphics[trim=49bp 114bp 18bp 107bp,clip,height=\pointmaprowheight]{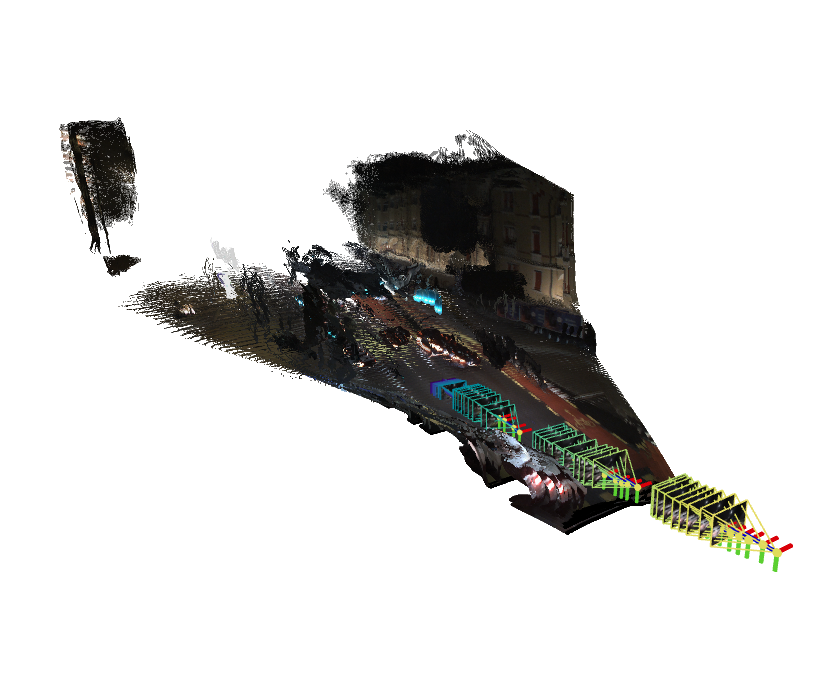}}}%
  {{\includegraphics[trim=128bp 87bp 136bp 153bp,clip,height=\pointmaprowheight]{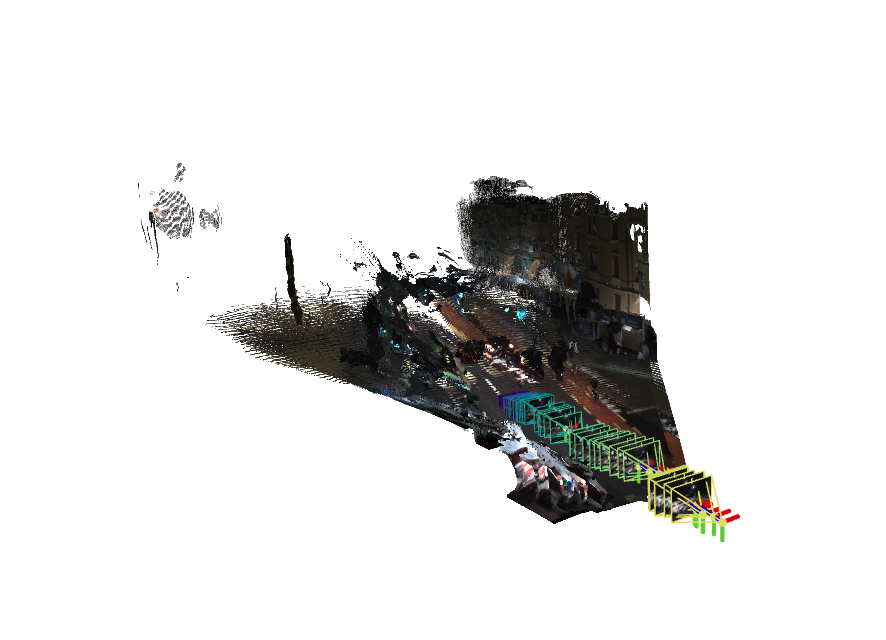}}}%
  {{\includegraphics[trim=107bp 23bp 53bp 25bp,clip,height=\pointmaprowheight]{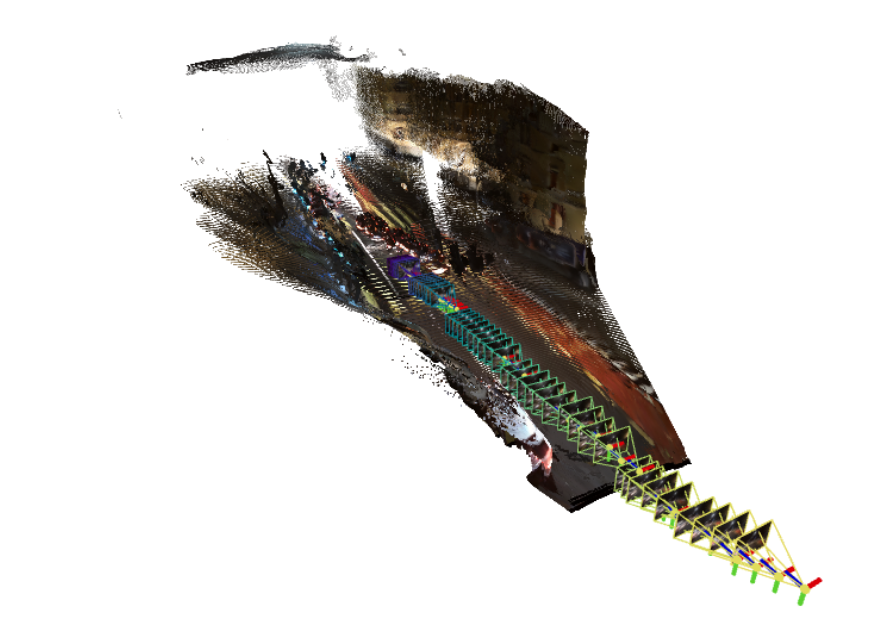}}}%
  {{\includegraphics[trim=204bp 82bp 133bp 90bp,clip,height=\pointmaprowheight]{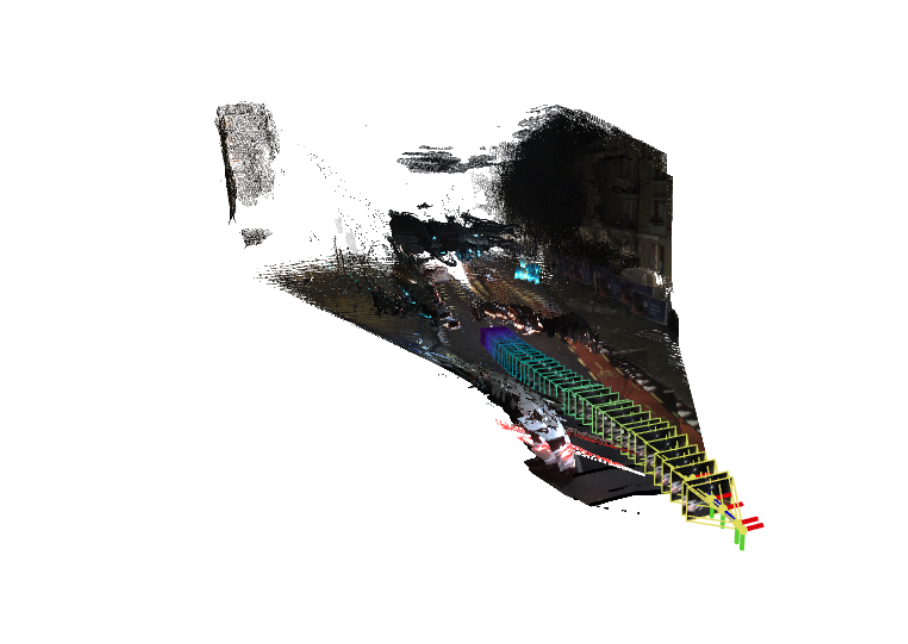}}}%
  {fig:pointmaps:dsec09}

\endgroup

\clearpage
\bibliographystyle{iclr2027_conference}
\bibliography{all,main}

\end{document}